\documentclass{bmvc2k}

\usepackage[dvipsnames]{xcolor}
\usepackage{epigraph}
\colorlet{revisioncolor}{black}
\newcommand{\rev}[1]{{\color{revisioncolor}#1}}
\hypersetup{
  colorlinks=true,
  linkcolor=MidnightBlue,
  citecolor=OliveGreen,
  urlcolor=BrickRed
}
\usepackage{multirow}

\title{An Exploratory Study on Abstract Images\\ and Visual Representations\\ Learned from Them}

\addauthor{Haotian Li}{hxl226@student.bham.ac.uk}{1}
\addauthor{Jianbo Jiao}{j.jiao@bham.ac.uk}{1}

\addinstitution{The MIx Group, School of Computer Science,
 University of Birmingham\\
 Birmingham, UK
}

\runninghead{Haotian Li, Jianbo Jiao}{An Exploratory Study on Abstract Images}

\def\eg{\emph{e.g}\bmvaOneDot}

\begin{document}

\maketitle

\begin{abstract}
Imagine living in a world composed solely of primitive shapes, could you still recognise familiar objects? Recent studies have shown that abstract images—constructed by primitive shapes—can indeed convey visual semantic information to deep learning models. However, representations obtained from such images often fall short compared to those derived from traditional raster images. In this paper, we study the reasons behind this performance gap and investigate how much high-level semantic content can be captured at different abstraction levels. To this end, we introduce the \textbf{H}ierarchical \textbf{A}bstraction \textbf{I}mage \textbf{D}ataset (HAID), a novel data collection that comprises abstract images generated from normal raster images at multiple levels of abstraction. We then train and evaluate conventional vision systems on \textit{HAID} across various tasks including classification, segmentation, and object detection, providing a comprehensive study between rasterised and abstract image representations. We also discuss if the abstract image can be considered as a potentially effective format for conveying visual semantic information and contributing to vision tasks. 
Project page: \url{https://fronik-lihaotian.github.io/HAID_page/}.
% Code and models will be made publicly available.\vspace{-4mm}

\end{abstract}\vspace{-4mm}

\epigraph{\makebox[0.9\textwidth][c]{\hspace{-12mm}``Art is the elimination of the unnecessary.''}}{\textit{--- Pablo Picasso}}\vspace{-5mm}

%-------------------------------------------------------------------------
\section{Introduction}\vspace{-2mm}
\label{sec:introduction}

\begin{figure*}[ht]
\centering
% \vspace{-2mm}
    \includegraphics[width=0.9\linewidth]{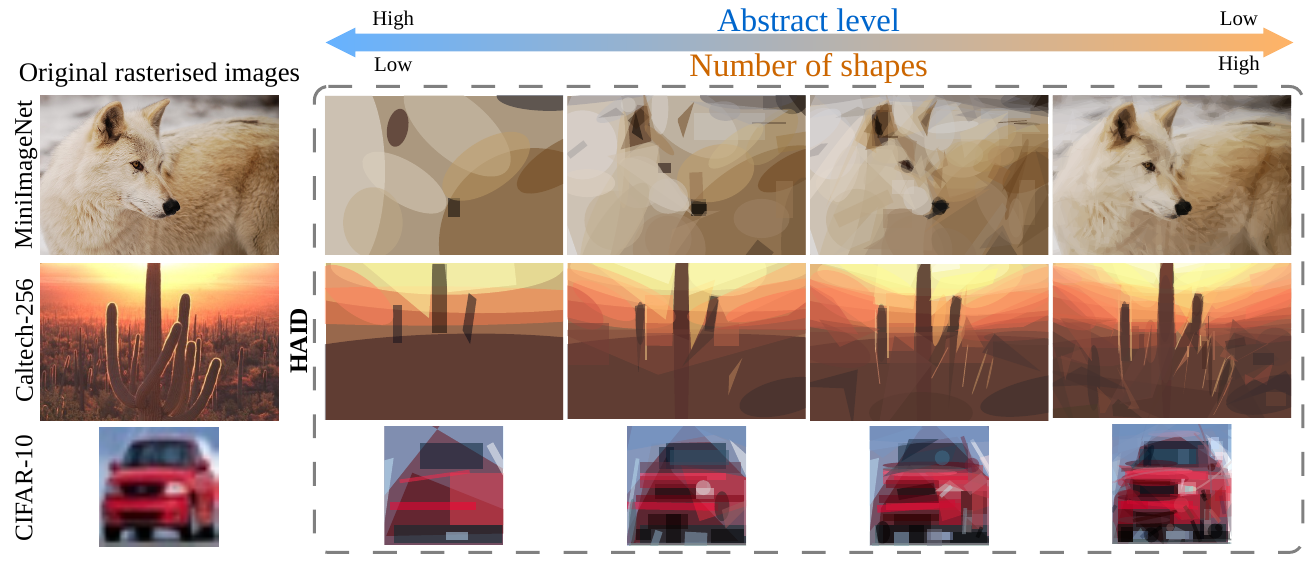}
    \vspace{-2mm}
    \caption{An overview of samples in the introduced HAID dataset and corresponding images from the raster image datasets. HAID-MiniImageNet supports the abstract level up to 1,000 shapes, HAID-Caltech-256 and HAID-CIFAR-10 support the abstract level up to 100 shapes.}\vspace{-5mm}
    \label{fig:title_pic}
\end{figure*}

Visual components, such as primitive shapes, are vital for humans to recognise and remember objects. Infants can classify objects based on their shapes \cite{infant_shape, infant_shape2, infant_shape3}, 
% humans can recognise objects merely based on shape information at little cost \cite{psy1}, 
and such shape cues can be quickly and efficiently extracted by the human brain \cite{psy2}. As for computer vision, abstract images are generally considered as the carrier to present the shape-oriented visual information. They are typically formed by vectorised shapes to provide lossless scalability and are widely used in many scenarios due to this special property. Although shape information plays a crucial role in human visual recognition patterns, early machine learning visual tasks did not focus too much on abstract images. Nevertheless, with the rapid development of computer vision, the potential contributions of such abstract images to machine learning systems are gradually being recognised. Remarkable progress related to vectorised image generation and understanding has been achieved, for example, DeepSVG \cite{carlier2020deepsvg} successfully generated the transition animation between two Scalable Vector Graphics (SVG) icons, SVGformer \cite{cao2023svgformer} further improved the performance and supported up to four different downstream tasks, recently, StarVector \cite{starvector} presented the first large-scale pretraining dataset and the Multi-modal Large Language Model (MLLM) for SVG generation. 

However, despite such great achievements, the studies related to abstract images generally stay on the high-abstract level and rarely consider the correlations with the complex visual semantic information from the real world. In the work of \cite{cai2023leveraging}, the authors try to leverage the powerful understanding abilities of the Large Language Model (LLM) to `see' and `draw' the vectorised images, but there is still a significant performance gap compared with the vision experts trained on pixel-level images. Another work \cite{sharma2024vision} also tried to use LLM to understand and generate code-based abstract images, then use the generated images to train the vision model and evaluate based on the real images. The result shows that LLM can understand and generate visual concepts from code-based images, yet it will fail when encountering images containing complex semantic information, and the contribution from generated images to vision systems is still limited. 

We are interested in the reason behind such performance gap. Following conclusions from some works \cite{cai2023leveraging, sharma2024vision}, we speculate that the difficulty of demonstrating the complex and fine-grained features from abstract images might be the major reason. In existing vector graphics image datasets \cite{SVGVAE, zou2024vgbench, carlier2020deepsvg, starvector}, most of them consist of simple and single-object icons or fonts. However, raster images generally have complex scenes with multiple objects. In the work of \cite{cai2023leveraging}, they tried to provide SVG images directly converted from rasterised images, but some fine-grained features, textural features, for example, still failed to be presented. Motivated by this, we are interested in asking: \textit{1) Is the level of abstraction a major reason for the performance gap between representations learned from raster and abstract images? 2) To what extent do changes in fine-grained features of abstract images affect the visual recognition of semantics?}

To answer these questions, we introduce a dataset called \emph{\textbf{H}ierarchical \textbf{A}bstraction \textbf{I}mage \textbf{D}ataset (HAID)} containing various abstract levels of SVG images. The dataset is generated directly from raster image datasets \cite{miniImagenet, caltech-256, cifar10} using the Primitive tool \cite{fogleman_primitive}. Then, we use images with different levels of abstraction to train and evaluate models for classification, object detection, and segmentation tasks. Finally, we discuss whether the difficulty of presenting fine-grained features is a major reason leading to such a performance gap and whether the abstract images generated by primitive shapes could contribute to the vision tasks.
To summarise, the main contributions of this study include:
\begin{itemize}[noitemsep, topsep=0pt]
  \setlength\itemsep{0.1em}     % space between items
    % \vspace{-2mm}
    \item We introduce a new dataset -- \emph{HAID} that comprises the different abstract levels of vectorised images generated from raster images.
    % \vspace{-2mm}
    \item A comprehensive study is presented on how the abstract level of images affects the ability of conventional vision systems to capture visual semantic information.
    % \vspace{-2mm}
    \item We investigate how much the abstract image representations from different abstract levels contribute to downstream tasks. 
    % \vspace{-2mm}
    \item We further discuss the potential benefits of abstract images to contribute to the vision tasks as well as the limitations
\end{itemize}\vspace{-4mm}

\begin{figure*}
    \centering
    \includegraphics[width=\linewidth]{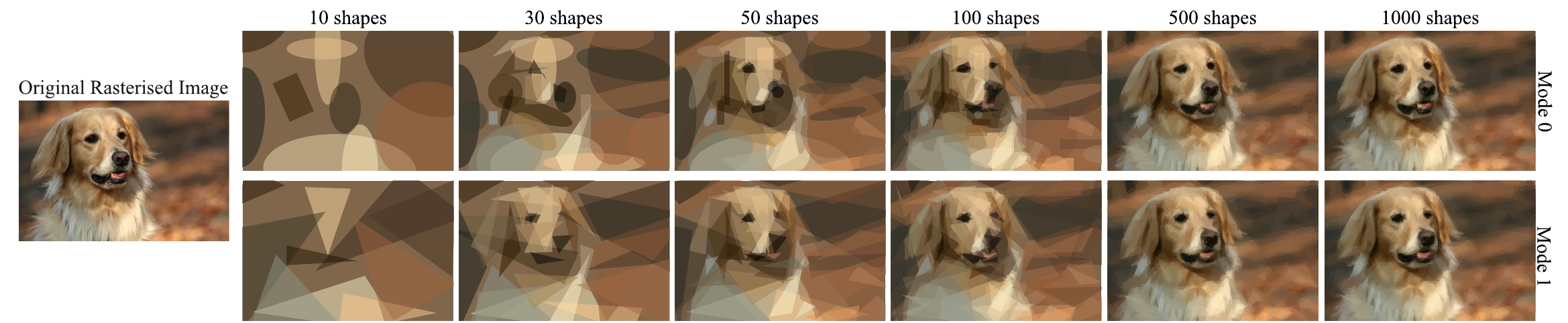}
    \vspace{-2mm}
    \caption{Sample from MiniImageNet (original rasterised image) and HAID-MiniImageNet (abstract images from different abstract levels ranging from 10 to 1,000 shapes).}\vspace{-5mm}
    \label{fig:HAIDexample_img}
\end{figure*}

% \begin{figure*}
%     \centering
%     \includegraphics[width=\linewidth]{pic/types_compar.drawio.pdf}
%     \vspace{-2mm}
%     \caption{Comparison between original raster image and corresponding abstract images generated with different type of primitive shapes. All abstract images share the same abstract level of 100 shapes.}\vspace{-2mm}
%     \label{fig:shape_type_compare}
% \end{figure*}

\section{Related Works}\vspace{-2mm}
\label{sec:related_work}

\paragraph{Abstract images.}

Abstract images, commonly rendered using vector graphics formats such as SVG \cite{SVG} and TikZ \cite{tikz}, have found widespread use in numerous domains due to their unique properties. Distinct from raster images, vector graphics offer lossless and infinite scalability, moreover, they are text-based, which facilitates both generation and subsequent editing. Scalable Vector Graphics (SVG) image, an XML-based format, is primarily considered in this study due to its convenience of third-party support and demonstration. Although abstract images enjoy certain advantages, their use mostly remains in presenting simple abstract information or logical relationships instead of fine features as high-resolution rasterised images.\vspace{-4mm}

\paragraph{Representation learning of vector graphics and datasets.}%\vspace{-2mm}

The previous deep learning studies on vector graphics and their datasets generally focused on the generation task \cite{SVGVAE, cao2023svgformer, carlier2020deepsvg, im2vec, vectorfusion, zou2024vgbench, starvector}, the early seminal works like SVG-VAE \cite{SVGVAE} and DeepSVG \cite{carlier2020deepsvg}, pioneered the ways of generating SVG images, and built the datasets that contained SVG fonts and icons. Subsequent studies further advanced these methods to improve performance as well as to broaden the scope of tasks related to vectorised images. For instance, by distilling from the powerful diffusion models, VectorFusion \cite{vectorfusion} is capable of directly generating SVG images from text instructions. Further, VGbench \cite{zou2024vgbench} leverages the Large Language Model (LLM) to endow the model with both visual understanding and generating abilities for vector graphics. The datasets for both works are formed as image-text pairs collected from past works or the Internet. Very recently, StarVector \cite{starvector} presented a foundation model for SVG generation as well as a new large-scale dataset. However, the datasets above are mostly built for the universal utilisation of vectorised graphics, and their images are often single-object and high-abstract which is disconnected from reality. 

With the remarkable progress achieved by LLMs, some studies try to utilise the textual property of vector graphics to endow the visual understanding ability to the LLM. Work \cite{cai2023leveraging} treats the SVG images as the bridge between image-text and enables the LLM in a variety of visual semantic understanding tasks. Moreover, the work \cite{sharma2024vision} tries to use code-based images to reveal whether the LLM can `see' and `draw'. Further, they use the code-based images drawn by LLM to train the vision system, which eventually demonstrates the ability to understand high-level visual semantic information from raster images. Despite all these studies exhibiting that abstract images can provide the visual semantic information for representation learning, a distinct performance gap persists between representations derived from pixel-level images and those obtained from abstract, code-based images. \vspace{-4mm}

\paragraph{Primitive.} %\vspace{-2mm}

Primitive \cite{fogleman_primitive} is a tool to generate abstract images from raster images. Different from VTracer \cite{vtracer2024} or Potrace \cite{potrace2003}, Primitive iteratively adds primitive shapes to a canvas to approximate the original raster image. Specifically, the algorithm randomly generates candidate primitive shapes at each iteration and then uses a hill-climb-based algorithm to repeatedly mutate these shapes, choosing the one with the best score evaluated by Root Mean Square Error (RMSE) as the target shape to be added to the canvas. The number of iterations is equivalent to the target number of shapes of abstract images. \rev{Due to the file capacity concern, Primitive is considered rather than VTracer. Potrace is excluded from consideration because it supports only binarised inputs (\eg black-and-white bitmaps), which do not meet our requirement for generating images from full-colour pixel inputs.} The comparison between Primitive and VTracer is shown in \cref{sec:supp_primtive_vs_vtracer}, and details of how the Primitive generates the shape-based images are shown in \cref{fig:Primitive_procedure} of the supplementary material. 
As Primitive can set different numbers and types of SVG primitive shapes to generate the target images, it is particularly well-suited in this project for simulating different abstraction levels in vectorised images. The effect of Primitive can be viewed in \cref{fig:HAIDexample_img}.\vspace{-4mm}

\section{Dataset}\vspace{-2mm}
\label{sec:Dataset}

To better discuss the questions mentioned above, here we introduce  a new dataset: \emph{\textbf{H}ierarchical \textbf{A}bstraction \textbf{I}mage \textbf{D}ataset (HAID)}, which comprises SVG images generated at multiple levels of abstraction from existing raster-image datasets, using the Primitive tool \cite{fogleman_primitive}. Specifically, the number of shapes determines the fine-grained level of the SVG images; as the number of shapes increases, the depicted objects as well as fine-grained details become increasingly recognisable from human perception (see \cref{fig:HAIDexample_img}). The dataset offers two primary advantages: \emph{(1) a one-to-one correspondence between the SVG image and their raster image counterparts, and (2) multiple abstraction levels for each raster image.} We analyse the differences between representations learned from pixel-level images and corresponding SVG images on three standard computer vision tasks: image classification, semantic segmentation, and object detection. For convenience, we term the HAID subset corresponding to a specific raster-image dataset as HAID-(name of the dataset), \eg HAID-MiniImageNet.\vspace{-4mm}

\subsection{Classification}\vspace{-2mm}

To obtain representations of code-based images and compare them with those derived from raster images, we generate SVG images from three open-source datasets. In this study, we primarily consider three datasets: MiniImageNet \cite{miniImagenet}, Caltech-256 \cite{caltech-256}, and CIFAR-10 \cite{cifar10}. An overview of the sample images generated from three datasets is shown in \cref{fig:title_pic}.

HAID-MiniImageNet is generated from MiniImageNet \cite{miniImagenet} using Primitive \cite{fogleman_primitive} with various numbers of shapes to simulate different levels of abstraction, ranging from 10 to 1,000 shapes \rev{(more details about the level split please refer to the \cref{sec:supp_shapes_config} of supplementary material)}. For each abstract level, similar to MiniImageNet, HAID-MiniImageNet contains 60,000 images across 100 categories. Sample SVG images are presented in \cref{fig:HAIDexample_img}. The two datasets are divided into training, validation, and testing sets in an 8:1:1 ratio in the same way. We select two options to generate the images by Primitive, using all types of shapes (mode 0) and using triangle shapes (mode 1). The reason we additionally generate the images constructed by triangles only is that this type of image shows the lowest capacity, which could be a potentially efficient form of abstract images. 

HAID-Caltech-256 is generated from Caltech-256 \cite{caltech-256} to support the classification task. The abstract levels of HAID-Caltech-256 range from 10 to 100 shapes. Both the original and abstract datasets are partitioned into training and validation sets using a 9:1 ratio. 

To comprehensively investigate the effect of image complexity, HAID-CIFAR-10, which is generated from CIFAR-10 \cite{cifar10} and characterised by comparatively simple images, is included, with the abstract levels similarly set between 10 and 100 shapes. The dataset splitting follows the official strategy.\vspace{-4mm}

\subsection{Object detection \& segmentation}\vspace{-2mm}

We further investigate whether the representations can contribute to other vision downstream tasks \eg semantic segmentation and object detection. Pascal VOC 2012 \cite{pascal-voc-2012} is used for these tasks. Following the official data split, the segmentation task utilises 1,464 images for training and 1,449 images for validation, while the object detection task utilises 5,717 training images and 5,823 validation images.\vspace{-4mm}

\section{Study on Abstract Images and Learned Representation}\vspace{-2mm}
\label{sec:experiments}

\begin{wrapfigure}{r}{0.5\textwidth}
    \centering
    \vspace{-4mm}\hspace{-4mm}
    \includegraphics[width=\linewidth]{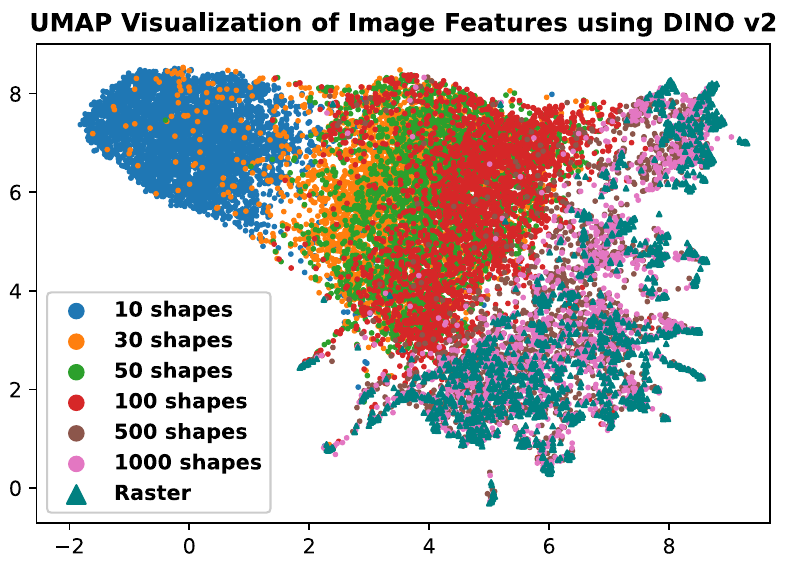}
    \vspace{-4mm}
    \caption{Visualisation of the difference across the abstract images and raster images in the embedding space by DINO v2.}\vspace{-5mm}
    \label{fig:umap}
\end{wrapfigure}

To comprehensively explore the issues outlined in the \cref{sec:introduction}, we first employ a third-party pretrained model to compare the difference between abstract and raster images, then, we evaluate model performance on HAID across three tasks: image classification, semantic segmentation, and object detection. In the classification task, we investigate whether traditional vision architectures can capture high-level semantic information from abstract images and how performance varies across different levels of abstraction. Subsequently, we employ the representations learned from the classification task as backbones and fine-tune the models for downstream tasks, assessing their contribution to enhanced performance in segmentation and object detection tasks.\vspace{-4mm}

\subsection{Difference across the abstract levels}\vspace{-2mm}

First, we investigate how many differences there are between abstract levels in the perspective of deep learning representations. We randomly sampled 4,000 image pairs—each consisting of an original MiniImageNet raster image and its corresponding abstract SVG versions across all abstract levels. We encoded these images using the DINO v2 \cite{dinov2} and applied UMAP \cite{umap} to project the resulting embeddings into two dimensions for visualisation, as shown in \cref{fig:umap}. From the visualisation results, embeddings of highly abstract images (\eg 10 or 30 shapes) form clusters that lie far from the raster‑image cluster, indicating substantial representational differences at coarse abstraction. As the number of primitives increases, the abstract‑image clusters move steadily closer to the raster‑image cluster, demonstrating that higher‑shape‑count abstractions maintain higher fidelity of semantics than lower ones. By the 500–1,000 shape levels, abstract embeddings overlap significantly with raster embeddings, suggesting near‑parity in semantic content despite the vectorised input format.

This trend confirms our intuition: coarse abstractions omit fine details and are thus distinct from pixel‑based representations, however, increasing the granularity of primitive shapes progressively bridges the gap. Consequently, the ``distance'' in embedding space represents the semantic fidelity of abstracted images relative to their raster counterparts. \vspace{-4mm}

\subsection{How significant are fine-grained features?}\vspace{-2mm}

Building on the previous visualisation, we next investigate how increasing abstraction—and the corresponding loss of fine‑grained details—impacts visual representation learning. To provide extensive studies of these issues, we decompose the problems into two sub-questions: 1) What is the performance of representation learning, and how does it compare to that achieved using raster images? 2) Will the performance be more comparable based on the low-resolution raster images that are difficult to display the fine-grained features? To answer these questions, we design a series of classification tasks to provide a comprehensive discussion. We primarily consider two conventional vision systems, ResNet50 \cite{resnet} and MobileNetv2 \cite{mobilenetv2}, to extract the semantic features. Our experiments utilise HAID-MiniImageNet, HAID-Caltech-256, and HAID-CIFAR-10, alongside their corresponding raster datasets, to enable a full comparison. All experiments are implemented in PyTorch and executed on an NVIDIA A100 40GB GPU.\vspace{-4mm}

\paragraph{Comparing with raster images, how good can it be?}
\label{sec:miniImage_test}

We discuss the difference between representations derived from raster images and those obtained from abstract images in this part. First, we establish baseline performance by training ResNet50  and MobileNetv2  on the original MiniImageNet dataset. Next, we train the models with the same architectures on the HAID-MiniImageNet dataset across six abstraction levels (10, 30, 50, 100, 500, and 1,000 shapes) and assess the performance on test sets corresponding to each level. Additionally, to examine the effect of training data volume on representation quality, we randomly select four subsets containing fewer training samples from the training set of HAID-MiniImageNet, ranging from 20$\%$ to 80$\%$ of the full training set. 

\begin{figure*}[!ht]
% \vspace{-3mm}
    \begin{minipage}[t]{0.5\linewidth}
        \centering
        \includegraphics[width=\textwidth]{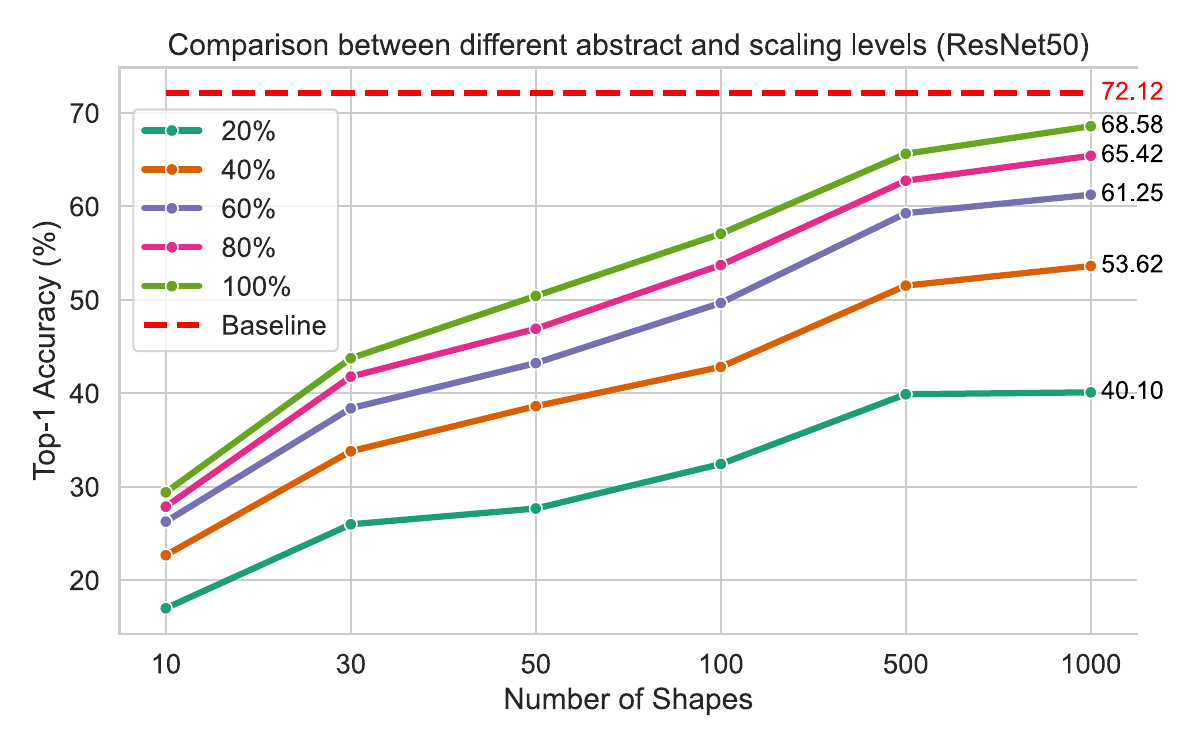}\vspace{-2mm}
        % \centerline{(a). ResNet50}
        \vspace{-2mm}
        \label{fig:test_acc_resnet_abs}
    \end{minipage}%
    \begin{minipage}[t]{0.5\linewidth}
        \centering
        \includegraphics[width=\textwidth]{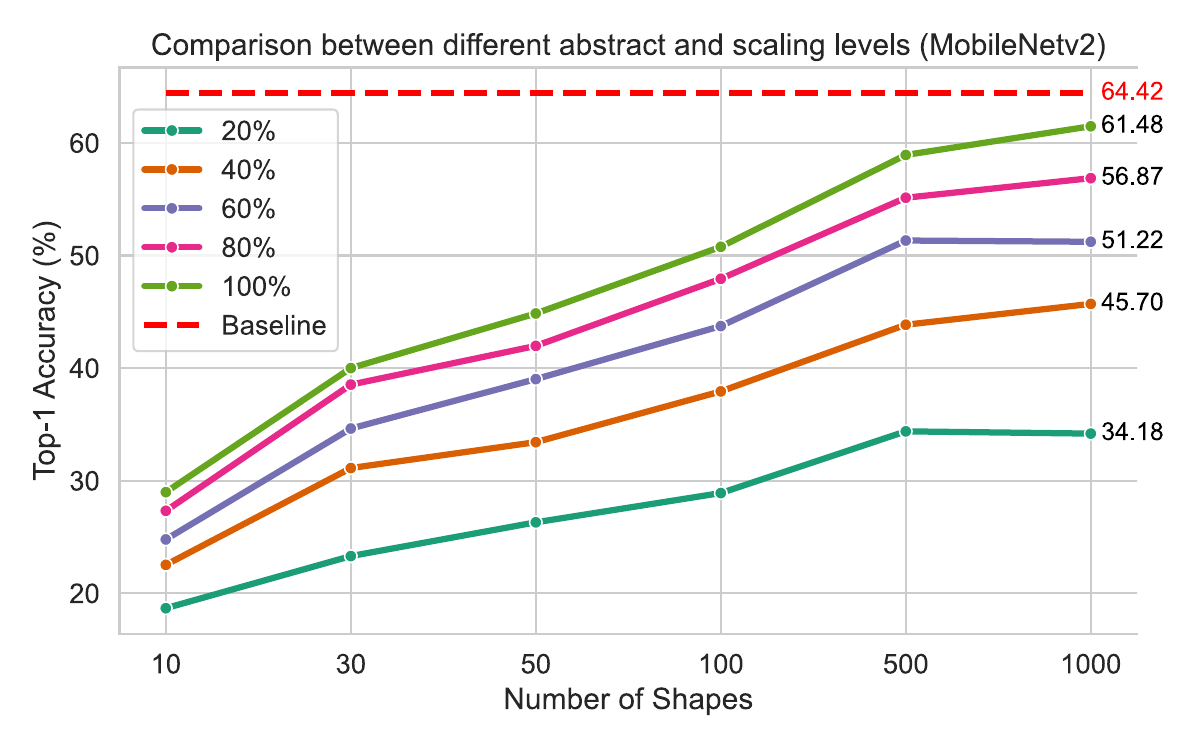}\vspace{-2mm}
        % \centerline{(b). MobileNet v2}
        \vspace{-2mm}
        \label{fig:test_acc_mv2_abs}
    \end{minipage}
    \vspace{-5mm}
    \caption{Comparison between representations learned from MiniImageNet and HAID-MiniImageNet across abstract levels (10--1,000 shapes) and scaling factors (20\%--100\%).}\vspace{-2mm}
    \label{fig:test_acc_same_abs}
\end{figure*}

For the training recipe, since we try to evaluate the difference between raster and abstract images rather than explore the best performance, simple hyperparameter settings are applied in our experiments. We use AdamW \cite{adamW} as the optimiser with the initial learning rate of 0.0001, and set batch size to 256. We also considered the data augmentations for training (more details are in \cref{sec:supp_classifi} of supplementary material). The training recipe is shared across all the experiments in this section. The final results are presented in the \cref{fig:test_acc_same_abs}, and the specific results are shown in \cref{tab:classification_test_acc_mode0,tab:classification_mv2_acc_mode0} of the supplementary material.
% \vspace{-3mm}

Our results indicate that as the level of fine-grained detail in the abstract images increases, the learned representations are better able to understand high-level semantic information, with performance gradually approaching that of the raster image baseline. In particular, SVG images generated with 500 and 1,000 shapes yield representations that are highly comparable to those derived from raster images. Conversely, at high abstraction levels (\eg 10 and 30 shapes), a pronounced performance gap is observed, which is expected given the inherent difficulty in recognising highly abstracted images even for humans. Additionally, our scaling experiments reveal that increasing the number of training samples further enhances the performance of the learned representations. 

\begin{figure*}[!ht]
% \vspace{-3mm}
    \begin{minipage}[t]{0.5\linewidth}
        \centering
        \includegraphics[width=\textwidth]{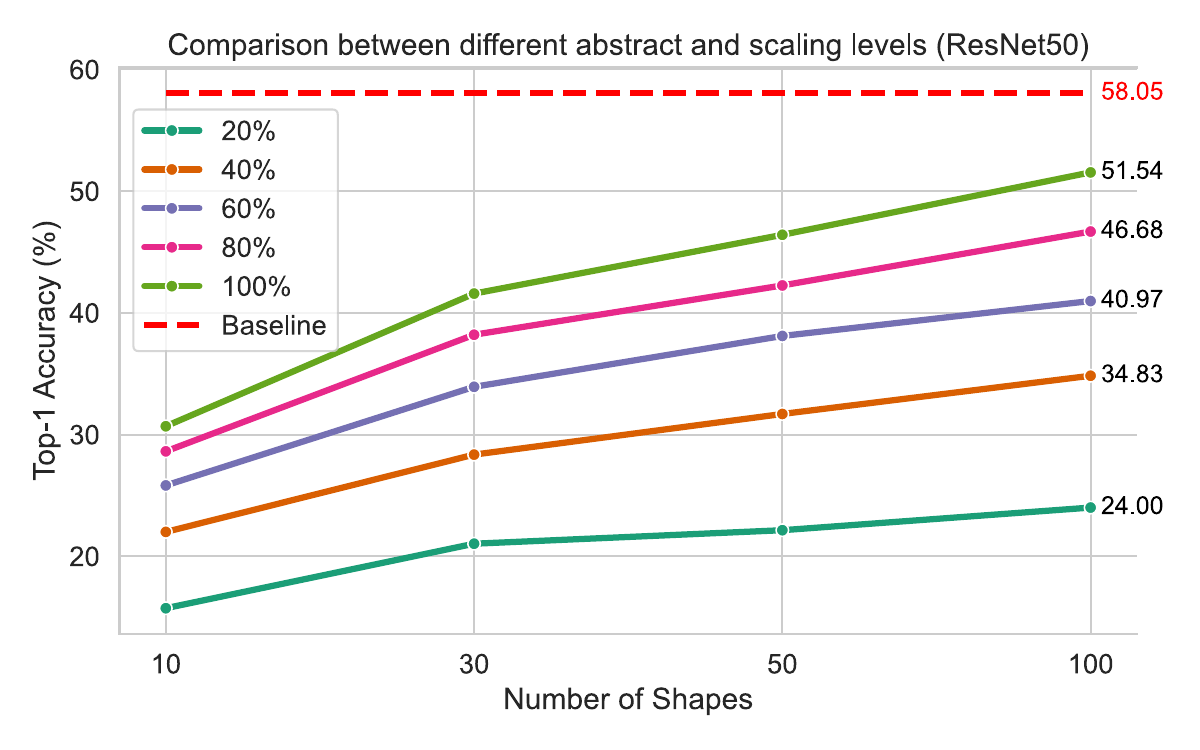}\vspace{-2mm}
        % \centerline{(a). ResNet50}
        \vspace{-4mm}
    \end{minipage}%
    \begin{minipage}[t]{0.5\linewidth}
        \centering
        \includegraphics[width=\textwidth]{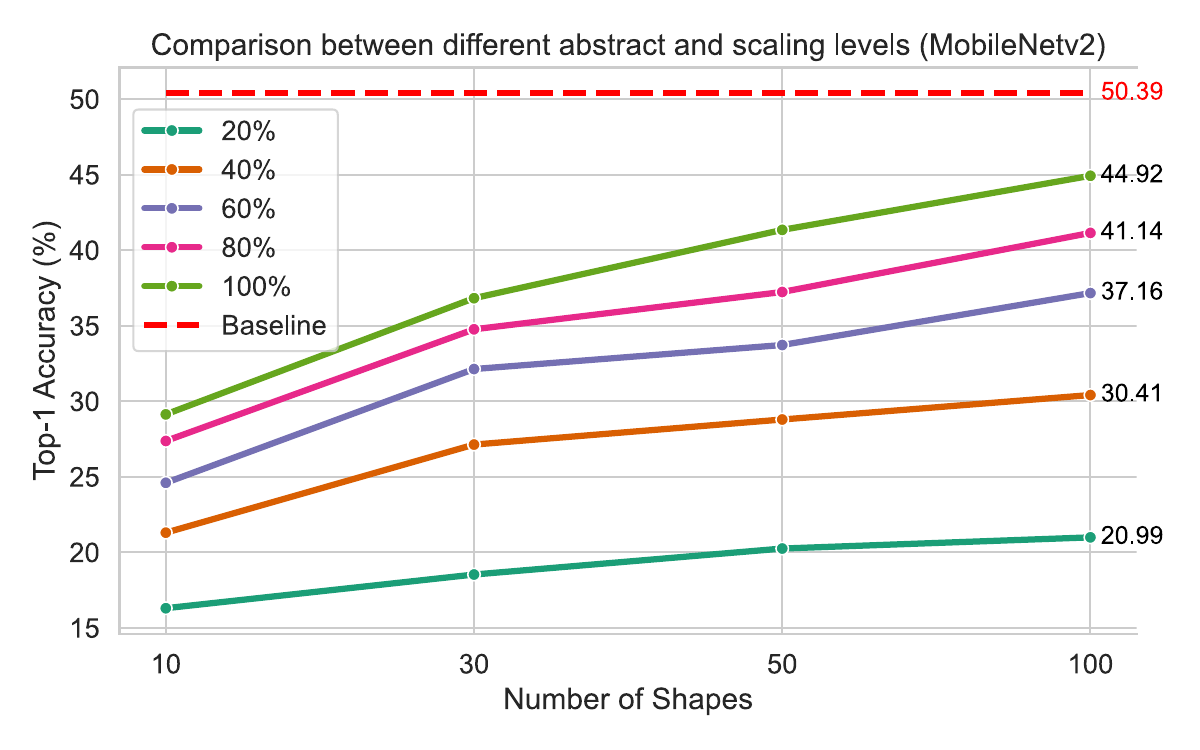}\vspace{-2mm}
        % \centerline{(b). MobileNet v2}
        \vspace{-4mm}
    \end{minipage}
    \vspace{-5mm}
    \caption{Comparison between representations learned from Caltech-256 and HAID-Caltech-256 across various abstract levels (10--100 shapes) and scaling factors (20\%--100\%).}\vspace{-5mm}
    \label{fig:test_acc_same_abs_cal}
\end{figure*}

We also evaluate the representation performance on another dataset to support our perspective. Similar to the experiments on HAID-MiniImageNet, we train ResNet50 and MobileNetv2 on the HAID-Caltech-256 across four abstraction levels (10, 30, 50, and 100 shapes). The baseline is built by training on the raster images of Caltech-256. The rest settings remain the same with the experiments on HAID-MiniImageNet. The results on HAID-Caltech-256 and Caltech-256, which are shown in \cref{fig:test_acc_same_abs_cal}, follow the trend of the results from HAID-MiniImageNet, demonstrating once again that the representations can better understand high-level semantics on images with low abstractions.

We also measured the difference between the abstractions with two different generation modes (abstractions with all types of shapes and triangles only). However, very slight differences are observed between them compared with the differences from abstract levels. So, the discussion regarding this part is narrated in \cref{sec:mode0vsmode1} of the supplementary material. \vspace{-4mm}

\begin{wraptable}{r}{0.65\textwidth}
    % \begin{table}[t]
    %     \centering
    \vspace{-5mm}
    \centering
    \caption{Accuracy on CIFAR-10 and HAID-CIFAR-10.}
    \label{tab:cifar-10}
    \small
    \setlength{\tabcolsep}{1mm}{
    \begin{tabular}{c|cccc|c} \toprule
        \textbf{Abstract level} & \textbf{10} & \textbf{30} & \textbf{50} & \textbf{100} & \textbf{Raster} \\ \midrule
        \textbf{Top-1 Acc} & 60.01\% & 67.02\% & 68.48\% & 70.17\% & 72.10\%  \\ \bottomrule
    \end{tabular}}
    \vspace{-4mm}
    % \end{table}
\end{wraptable}

\paragraph{Recognising small images.}

To further investigate whether the difficulty in demonstrating fine-grained details is the primary factor influencing performance, we employ the CIFAR-10 dataset \cite{cifar10}, which contains low-resolution images that are also difficult to present visual details. A four-layer convolutional neural network is used to extract features from the images. The network is trained from scratch on HAID-CIFAR-10 at various abstraction levels for 10 epochs (more training details are in the supplementary material), and the top-1 classification accuracy of the resulting representations is evaluated. The results are summarised in \cref{tab:cifar-10}.

Notably, although a small performance gap remains, the performance gap within highly abstracted levels area between raster images and abstract images is significantly reduced. Combined with previous experimental results, this observation suggests that the inability to capture fine-grained details is a major factor contributing to the performance gap between representations derived from raster images and those obtained from code-based images.\vspace{-4mm}

\subsection{Can the representations further contribute the downstream tasks?}\vspace{-2mm}

We further investigated how much these representations can contribute to downstream tasks by examining two benchmarks: semantic segmentation and object detection.\vspace{-4mm}

\begin{figure*}[!ht]
    \begin{minipage}[t]{0.5\linewidth}
        \centering
        \vspace{-4mm}
        \includegraphics[width=\textwidth]{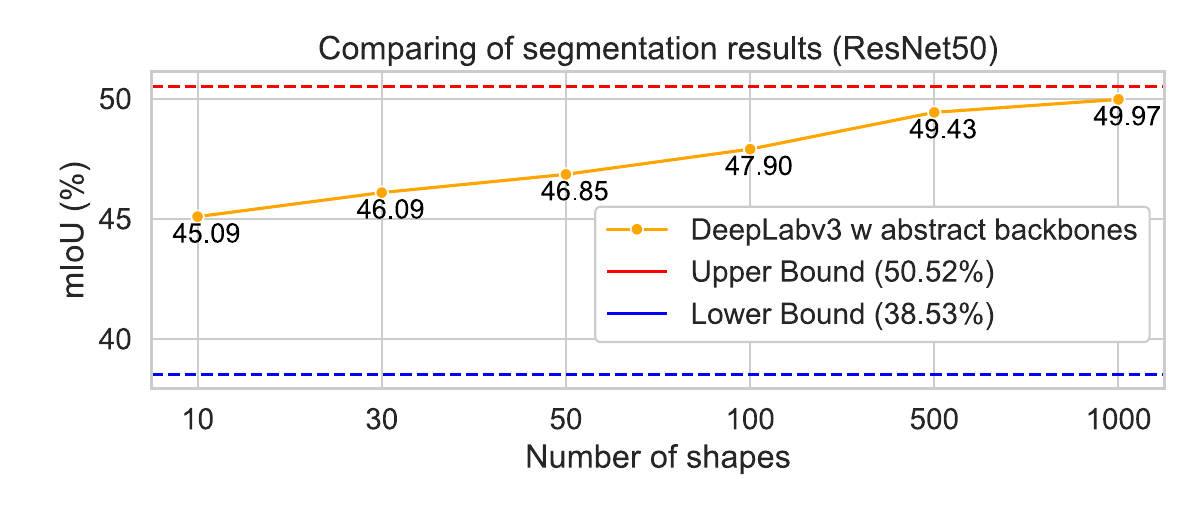}
        % \centerline{(a). DeepLabv3-ResNet50}
    \end{minipage}%
    \begin{minipage}[t]{0.5\linewidth}
        \centering
        \vspace{-4mm}
        \includegraphics[width=\textwidth]{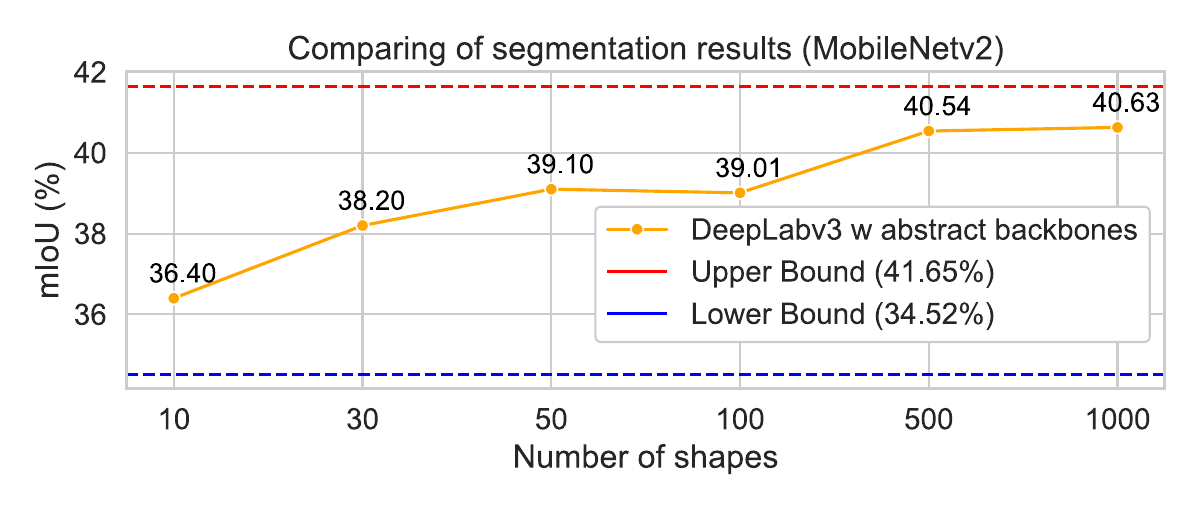}
        % \centerline{(b). DeepLabv3-MobileNet v2}
    \end{minipage}
    \vspace{-4mm}
    \caption{Semantic Segmentation results of DeepLabv3 with backbones and two baselines, upper bound refers to the model initialised with backbone pretrained on MiniImageNet, lower bound refers to the model with random initialisation.}\vspace{-4mm}
    \label{fig:segmentation_with_bb}
\end{figure*}

\begin{figure*}[!ht]
    \begin{minipage}[t]{0.5\linewidth}
        \centering
        \vspace{-4mm}
        \includegraphics[width=\textwidth]{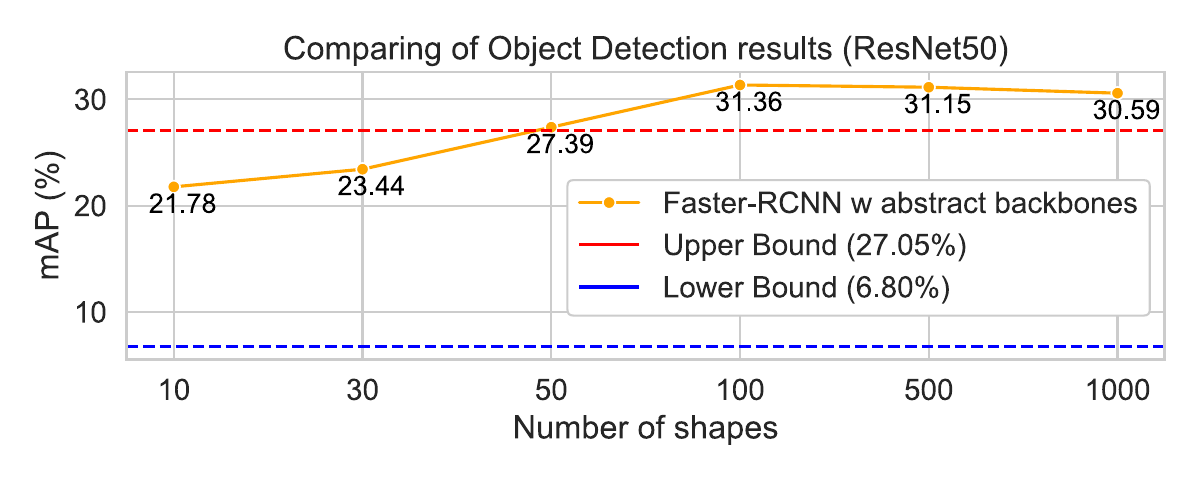}
        % \centerline{(a). Faster R-CNN}
    \end{minipage}%
    \begin{minipage}[t]{0.5\linewidth}
        \centering
        \vspace{-4mm}
        \includegraphics[width=\textwidth]{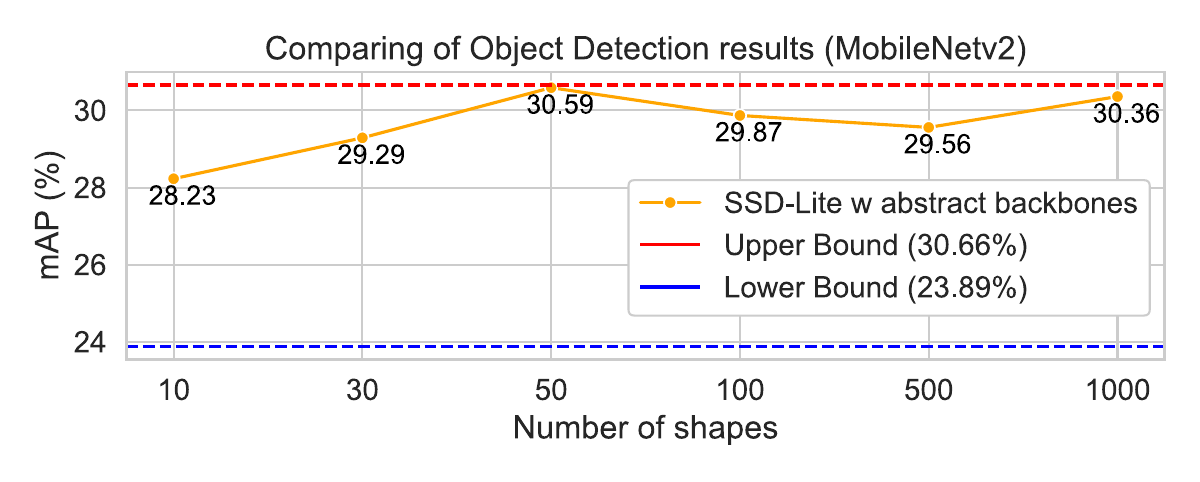}
        % \centerline{(b). SSD-Lite}
    \end{minipage}
    \vspace{-4mm}
    \caption{Object Detection results of Faster R-CNN and SSD-Lite with backbones and two baselines. The upper bound refers to the model initialised with a backbone pretrained on MiniImageNet, and the lower bound refers to the model with random initialisation.}\vspace{-2mm}
    \label{fig:object_detection_with_bb}
\end{figure*}

\paragraph{Semantic segmentation.}

To evaluate if the abstract image representations can contribute to the downstream task of semantic segmentation, we utilise backbones derived from models trained on both MiniImageNet and HAID-MiniImageNet (see \cref{sec:miniImage_test}). DeepLabv3 \cite{deeplabv3} is considered the framework for the segmentation tasks. For comprehensively measuring the contributions, we also set two performance baselines, one using the backbone pretrained on MiniImageNet, and another without any initialisation from pretrained backbone. 

The models are trained using the AdamW optimiser with an initial learning rate of 0.0009, batch size of 8, for 200 epochs. We compared the results of DeepLabv3 models with both ResNet50 and MobileNetv2 backbones in \cref{fig:segmentation_with_bb}. The numerical differences between different model performances and upper and lower bounds are presented in \cref{tab:segmentation}.

From the results, initialising the network weights from pretrained abstract backbones shows an increasing performance trend as the descending of abstraction level. Notably, regardless of the specific abstraction level employed, the contributions of these representations are evident, demonstrating that such representations can further contribute to segmentation tasks, even if such tasks strongly rely on fine-grained features capturing ability. \vspace{-4mm}

\paragraph{Object detection.}

After evaluating the results from the downstream task, which challenges the pixel-level visual understanding, we then discuss how abstract image representation contributes to the spatial visual understanding in the object detection task. Two architectures: SSD-Lite \cite{mobilenetv2} and Faster R-CNN \cite{fasterrcnn} are considered. For SSD-Lite—a lighter variant of SSD \cite{ssd}—we initialised the weights from MobileNetv2 backbones. The model was trained using the stochastic gradient descent (SGD) optimiser with an initial learning rate of 0.001 and a weight decay of 0.0005 for 120 epochs. In addition, we used Faster R-CNN with a ResNet50 backbone for object detection. This model was trained for 20 epochs using the SGD optimiser with an initial learning rate of 0.005 and a weight decay of 0.0005. The detail setting refers to \cref{sec:supp_seg_obj} of supplementary material. \Cref{fig:object_detection_with_bb} compares the difference between the model initialised by the abstract image backbone and two baselines. The specific numerical differences are shown in \cref{tab:object_detection} of the supplementary material.

The results from object detection demonstrate the same trend as the results from semantic segmentation. Moreover, the performance for some models initialised by representations from abstract images surprisingly exceeds the performance from the raster image representation. \rev{From the Grad-CAM \cite{grad-cam} visualisation of Faster R-CNN initialised by different backbones, we found an interesting phenomenon, that the attention map from the model with abstract prior concentrated more on core semantic area of the objects. Such effect is most pronounced at 100 shapes, but as the number of shapes further increases, this effect gradually disappears and approaches the model with raster prior (details are shown in \cref{sec:supp_seg_obj_results} of supplementary material).} These results once again exhibit that representations obtained from abstract images can contribute to downstream tasks, and more than that, compared with the results from previous segmentation tasks, we can observe that tasks relying on spatial perception, such as object detection, seem to better reflect the advantages of abstract image representation compared to tasks that rely more on fine-grained features. \vspace{-4mm}

\subsection{How human perceive abstract images?}\vspace{-2mm}
\label{sec:user_study}

\rev{
To further evaluate our dataset from a human perception perspective, we conducted a user study to quantify how confidently humans perceive object identity in HAID abstractions. The images with only 10-shape abstract level were excluded, as they are almost certainly unrecognisable. We chose 36 images from HAID-MiniImageNet and MiniImageNet at six levels (30, 50, 100, 500, 1,000 shapes, and original images). 
% All images were balanced by a priori difficulty, which is categorised into ``easy samples'' and ``hard samples''. 
Participants are asked to provide a 1-5 rating for each image to indicate how confident they are in recognising the object(s) within it. Detailed design of the user study is explained in \cref{sec:supp_user_study} of supplementary material.

\begin{figure*}[!ht]
\vspace{-3mm}
    \begin{minipage}[t]{0.5\linewidth}
        \centering
        % \vspace{-4mm}
        \includegraphics[width=\textwidth]{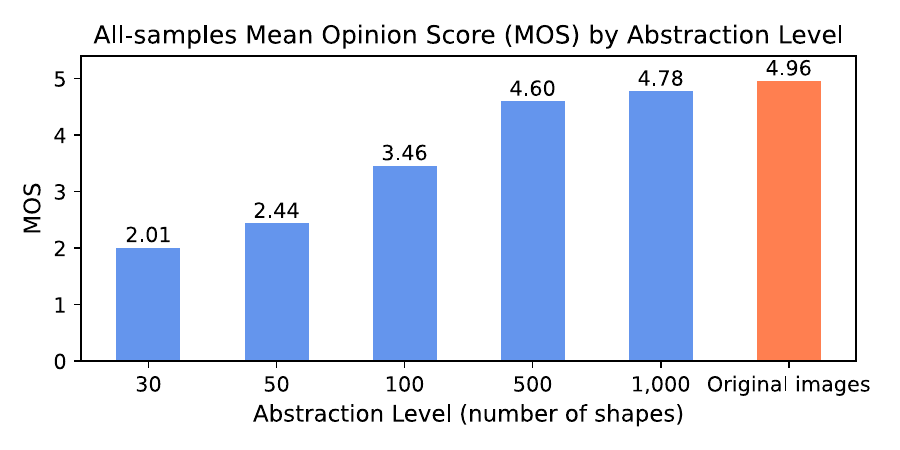}
    \end{minipage}%
    \begin{minipage}[t]{0.5\linewidth}
        \centering
        % \vspace{-4mm}
        \includegraphics[width=\textwidth]{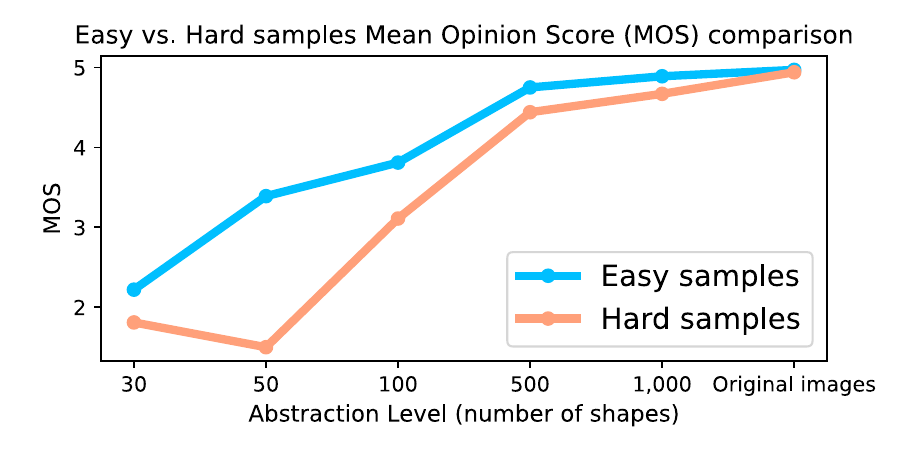}
    \end{minipage}
    \vspace{-6mm}
    \caption{Left chart demonstrates the MOS across the different abstract levels as well as original images; the right chart compares the differences between `hard' and `easy' samples.}\vspace{-4mm}
    \label{fig:user_study}
\end{figure*}

% TODO fill in the ref of the table and figures
We received responses from 12 participants in total and collected their Mean Opinion Score (MOS) of each abstraction level to produce three summary series: \textit{MOS for easy samples}, \textit{hard samples} (according to the complexity, more details in the supplementary material), and \textit{MOS for all samples}. The results are presented in \cref{fig:user_study} (comprehensive results in \cref{sec:supp_user_study} of supplementary material). We can see that MOS for all samples increases monotonically with the number of primitives. Easy samples show higher confidence scores at low shape counts (\eg 2.22 vs. 1.81 at 30 shapes), while hard samples require substantially more primitives before the score approaches that of the originals (notably the dip at 50 shapes for hard samples).

Some observations are derived from the above analysis: 1) HAID abstractions retain perceptually relevant structure: at moderate-to-high fidelity (500 and 1,000 primitives), observers report confidence of perception close to original images. 2) Harder samples require more primitives to reach comparable perceptual clarity, suggesting adaptive allocation of abstraction budget may benefit recognition tasks that require fine-grained discrimination.
}\vspace{-4mm}

\subsection{Potential benefits of abstract images}\vspace{-2mm}

In this section, we discuss whether the abstract images can be a potentially effective format to contribute to the vision tasks, as well as their limitations. As the previous results demonstrated, representations learned from sufficiently detailed abstract images (\eg those generated with 500 and 1,000 shapes) approach—and in some object‑detection scenarios even exceed—the performance of raster‑trained representations. Considering the pixel-level images take advantage of using CNN-based models, which are designed for the rasterised images, not abstract images, such results are very promising to further explore how the abstract images can contribute to the vision tasks. 

However, two key limitations remained. First, highly abstract images (fewer than 100 shapes) lack critical fine‑grained features—such as texture, small edges, or subtle shading—that raster data naturally provides. As a result, performance gaps persist in tasks heavily dependent on such details (\eg semantic segmentation). Second, in this study, we use Primitive to generate the abstract images that approximate the original raster images. Despite the resulting images being visually appealing, the redundant shapes may be introduced during the generation process, which leads to unnecessary code fields and increases the capacity of the file. In \cref{sec:entopy} of supplementary material, we observe a strong correlation between perceptual similarity and image entropy, which shows that under the same abstract level, images enjoying low entropy generally have better perceptual loss on related abstract images, in other words, entropy can be considered as the metric to provide the trade-off between capacity and performance to further improve the efficiency. Moreover, considering that abstract images enjoy code-based format, the keywords of the SVG code can be further compressed and thus benefiting data transmission. \vspace{-4mm}

\section{Conclusion}\vspace{-2mm}
\label{sec:conclusion}

In this paper, we investigated abstract images and performed a study on the representations learned from them. Experiments showed that fine-grained detail is one of the main factors leading to the performance gap between representations learned from raster images and those learned from abstract images. On the other hand, as the level of fine-grain increases, the capacity of representation to capture high-level semantic information improves, thereby narrowing such performance differences. Moreover, our downstream task experiments revealed that representations derived from abstract images can effectively contribute to visual tasks, even achieving comparable performance on tasks that are less reliant on fine-grained details. \rev{From the analysis we found that models initialised from backbones pretrained on abstract images show stronger feature attention to object geometry and contours, yielding improved bounding-box localisation in Faster R-CNN, with peak gains at moderate abstraction (100 shapes).} Given the inherent advantages- including lossless scalability, a compact textual format, and ease of editing the abstract images show significant promise as a novel data form for visual representation learning and related vision tasks. %\vspace{-2mm}

\section*{Acknowledgment}
This project is partially supported by the Royal Society grants (SIF\textbackslash R1\textbackslash231009, IES\textbackslash R3\textbackslash223050) and an Amazon Research Award.
The computations in this research were performed using the Baskerville Tier 2 HPC service. Baskerville was funded by the EPSRC and UKRI through the World Class Labs scheme (EP\textbackslash T022221\textbackslash1) and the Digital Research Infrastructure programme (EP\textbackslash W032244\textbackslash1) and is operated by Advanced Research Computing at the University of Birmingham.

\bibliography{egbib}

\clearpage

% \maketitlesupplementary
\setcounter{section}{0}
\setcounter{table}{0}
\setcounter{figure}{0}
\setcounter{page}{1}

\begin{center}
\textbf{\color[rgb]{0,.1,.4}{\Large An Exploratory Study on Abstract Images\\ and Visual Representations Learned from Them (Supplementary Material)}}
\end{center}

\renewcommand{\thefigure}{S\arabic{figure}}
\renewcommand{\thetable}{S\arabic{table}}
\renewcommand{\thesection}{S\arabic{section}}

\section{Dataset}
\label{sec:sm_dataset}

\subsection{More examples}
\begin{figure*}[!ht]
    \centering
    \includegraphics[width=0.8\linewidth]{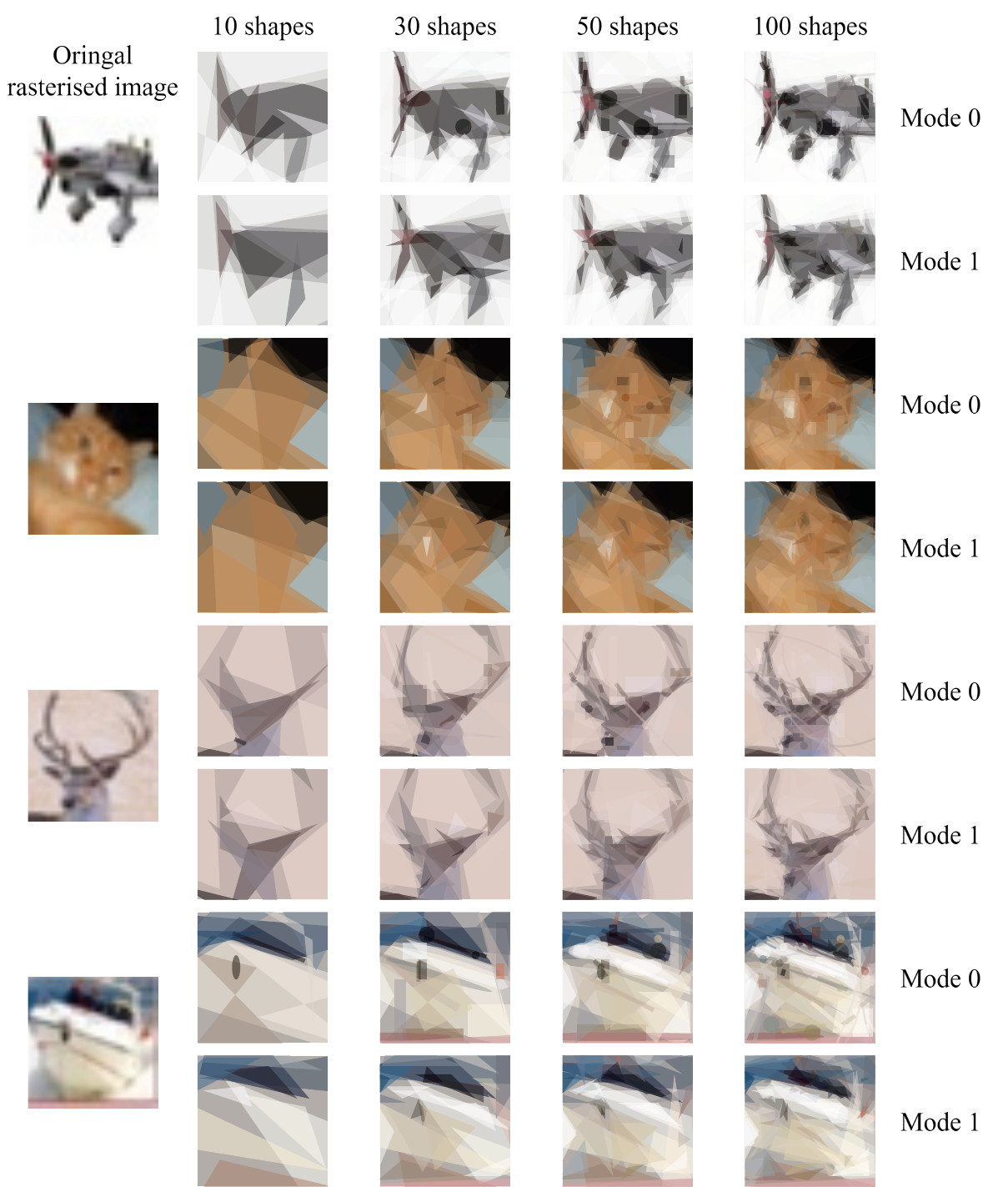}
    \caption{Example image pairs between CIFAR-10 and HAID-CIFAR-10}
    \label{fig:cifar_10_examples}
\end{figure*}

In here, we demonstrate more examples of our dataset. More image pairs of MiniImageNet and HAID-MiniImageNet are shown in \cref{fig:examples_of_HAID_MiniImageNet} and more image pairs of CIFAR-10 and HAID-CIFAR-10 are shown in \cref{fig:cifar_10_examples}.

\rev{
\subsection{Configurations of HAID}
\label{sec:supp_shapes_config}

HAID contains three sub-datasets generated from MiniImageNet \cite{miniImagenet}, Caltech-256 \cite{caltech-256}, CIFAR-10 \cite{cifar10}. We applied different abstract levels when generating these three datasets, specifically, HAID-CIFAR-10 and HAID-Caltech-256 support the SVG images with abstract levels up to 100 shapes; however, the HAID-MiniImageNet supports two extra abstract levels: 500 and 1,000 shapes. The reason for applying two more abstract levels for HAID-MiniImageNet is that they can provide comprehensive results about how primitive-based images perform on images with complex scenes. We observed that, compared with Caltech-256, MiniImageNet contains many more samples with multiple objects and complex backgrounds. As for CIFAR-10, generating images with shapes of 500 and 1,000 will far exceed the size of the original pixel image (\textasciitilde1 KB vs. \textasciitilde60 KB). We believe that, for this study, it is meaningless to compare the performance under significant file size differences. 

\subsection{Primitive \textit{vs.} VTracer}
\label{sec:supp_primtive_vs_vtracer}

VTracer \cite{vtracer2024} is a tracing-based algorithm that can convert pixel images to SVG images. Although it can faithfully restore the details of pixel images, the file size of the generated SVG images is generally much larger than the original pixel images. \Cref{fig:primitive_vs_Vtracer} compares the differences in generation effect and file size between images generated by the Primitive and VTracer algorithms. 

\begin{figure*}
    \centering
    \includegraphics[width=\linewidth]{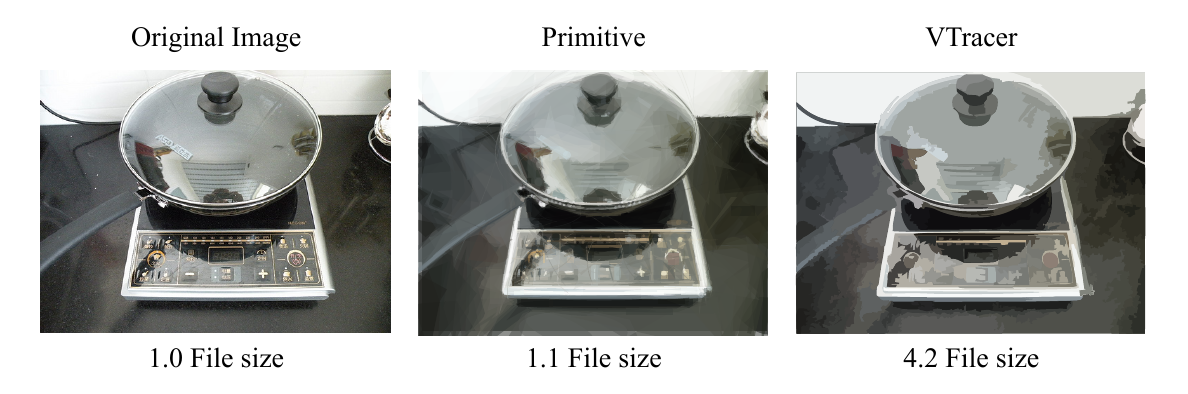}
    \caption{The comparison between the images generated by the Primitive and VTracer. We applied the highest capacity setting (1,000 shapes) to generate the Primitive-based SVG images. For the image generated by the VTracer, the hyperparameters will be: Filter Speckle = 15, Colour Precision = 6, and Gradient Step = 16. Based on the file size of the original image, we show the difference in file size between the generated images and their original image in the form of multiples.}
    \label{fig:primitive_vs_Vtracer}
\end{figure*}

}

% In \cref{sec:miniImage_test}, we discussed the visual different of abstract images between using all types of primitive shapes (mode 0) and triangle only (mode 1). The fine-grained difference of abstract images between mode 0 and mode 1 are shown in \cref{fig:mode0vsmode1}.

\section{Supplementary Experiment Content}
\label{sec:supp_experi_results}

\subsection{Experiment setting}

In here, we supplement the experimental details in \cref{sec:experiments}, including the settings of hyperparameters and network architectures. The parameter may not be the best setting since we focus more on comparing the performance difference under the same situations rather than exploring the best performance.

\paragraph{Classification.}
\label{sec:supp_classifi}

For the experiments of training ResNet50 \cite{resnet} and MobileNetv2 \cite{mobilenetv2} on the MiniImageNet and HAID-MiniImageNet, we use AdamW as the optimiser with an initial learning rate of 0.0001, and we use random resize crop to frame the image size into 224$\times$224. We use a series of data augmentation strategies, including: Random Horizontal Flipping, Random Augment \cite{randaugment}, and Random Erasing \cite{randomerasing}, training for 120 epochs. For the HAID-CIFAR-10 experiments, we only use the Adam \cite{adam} optimiser with an initial learning rate of 0.001 and training for 10 epochs.

\begin{table*}[!ht]
    \centering
    \caption{Top-1 Accuracy of ResNet 50 with all shapes (mode0) on MiniImageNet and HAID-MiniImageNet}
    \begin{tabular}{l|llllll|l}\toprule
        \multirow{2}{*}{Settings} & \multicolumn{6}{c|}{\textbf{Mode 0}} &  \\ \cmidrule{2-8}
        & \textbf{10} & \textbf{30} & \textbf{50} & \textbf{100} & \textbf{500} & \textbf{1,000} & \textbf{Raster} \\ \midrule
        \textbf{20\%} & 17.00\% & 25.98\% & 27.67\% & 32.43\% & 39.90\% & 40.10\% & 42.67\%  \\ 
        \textbf{40\%} & 22.67\% & 33.80\% & 38.62\% & 42.82\% & 51.52\% & 53.62\% & 55.78\%  \\ 
        \textbf{60\%} & 26.27\% & 38.40\% & 43.23\% & 49.67\% & 59.27\% & 61.25\% & 63.88\%  \\ 
        \textbf{80\%} & 27.87\% & 41.78\% & 46.90\% & 53.72\% & 62.75\% & 65.42\% & 67.92\%  \\ 
        \textbf{100\%} & 29.40\% & 43.75\% & 50.42\% & 57.07\% & 65.63\% & 68.58\% & 72.12\%  \\ \bottomrule
    \end{tabular}
    \label{tab:classification_test_acc_mode0}
\end{table*}

\begin{table*}[!ht]
    \centering
    \caption{Top-1 Accuracy of MobileNetv2 with all shapes (mode 0) on MiniImageNet and HAID-MiniImageNet}
    \begin{tabular}{l|llllll|l} \toprule
        \multirow{2}{*}{Settings} & \multicolumn{6}{c|}{\textbf{Mode 0}} &   \\ \cmidrule{2-8}
        & \textbf{10} & \textbf{30} & \textbf{50} & \textbf{100} & \textbf{500} & \textbf{1,000} & \textbf{Raster} \\ \midrule
        \textbf{20\%} & 18.67\% & 23.30\% & 26.30\% & 28.90\% & 34.38\% & 34.18\%  & 34.12\%  \\ 
        \textbf{40\%} & 22.52\% & 31.12\% & 33.42\% & 37.93\% & 43.85\% & 45.70\%  & 49.00\%  \\ 
        \textbf{60\%} & 24.78\% & 34.63\% & 39.02\% & 43.73\% & 51.33\% & 51.22\%  & 55.72\%  \\ 
        \textbf{80\%} & 27.32\% & 38.53\% & 41.97\% & 47.93\% & 55.13\% & 56.87\%  & 61.05\%  \\ 
        \textbf{100\%} & 28.98\% & 40.00\% & 44.85\% & 50.77\% & 58.93\% & 61.48\%  & 64.42\%  \\ \bottomrule
    \end{tabular}
    \label{tab:classification_mv2_acc_mode0}
\end{table*}

\paragraph{Semantic Segmentation.}
\label{sec:supp_seg_obj}

During training the DeepLabv3 for Semantic Segmentation, we randomly crop the image size to 480$\times$480 from the base size 520$\times$520. We use the AdamW optimiser with an initial learning rate of 0.0009, momentum of 0.9, and weight decay of 0.01 to train the models for 200 epochs. 

\paragraph{Object Detection.}
For Faster R-CNN, we use Stochastic Gradient Descent (SGD) as the optimiser with an initial learning rate of 0.005, momentum equals 0.9, and weight decay is 0.0005. The learning rate is reduced by a factor of 0.95 every 5 epochs, and we trained Faster R-CNN for 20 epochs. For the model architecture of Faster R-CNN, the Region Proposal Network (RPN) uses an anchor generator with scales of 32, 64, 128, 256, 512 and aspect ratios of 1:2, 1:1, 2:1. For the Region of Interest (ROI) Pooling, a single-scale of RoIAlign is employed with an output size of 7 and a sampling ratio of 2. 

For SSD-Lite, we use SGD as the optimiser with an initial learning rate of 0.001, momentum equals 0.9, and weight decay equals 0.0005. The learning rate was reduced by a factor of 0.1 at 80 and 100 epochs. The implementation detail follows \href{https://github.com/qfgaohao/pytorch-ssd}{this repository}. 

\subsection{Supplementary results}

\paragraph{Classification.} 

The full results from HAID-MiniImageNet are shown in \cref{tab:classification_test_acc_mode0} and \cref{tab:classification_mv2_acc_mode0}.

\paragraph{Semantic Segmentation \& Object Detection.}
\label{sec:supp_seg_obj_results}

In \cref{sec:experiments}, we showed the experiment results in \cref{fig:segmentation_with_bb} and \cref{fig:object_detection_with_bb}. In here, we further demonstrate the full comparison of performance between models initialised by the backbone pretrained on HAID-MiniImageNet and two baselines (Upper Bound and Lower Bound) on two downstream tasks. \Cref{tab:segmentation} shows the results from semantic segmentation task and \cref{tab:object_detection} shows the results from object detection task.

\begin{table*}[!ht]
    \centering
    \caption{Comparing the performances by Mean Intersection over Union (mIoU) of DeepLabv3 with backbones of ResNet 50 (up) and MobileNetv2 (down) from MiniImageNet and HAID-MiniImageNet. The number in the first line represents the abstract level applied for backbones, from 10 to 1,000 shapes. We also provide the upper bound (UB: fine-tuning with backbone from raster images) and lower bound (LB: training without backbone), and compare them with fine-tuning models with backbones gained from HAID-MiniImageNet.}
    \begin{tabular}{c|cccccc} \toprule
        ~ & 10 & 30 & 50 & 100 & 500 & 1,000 \\ \midrule
        DeepLabv3-ResNet50 & 45.09 & 46.09 & 46.85 & 47.90 & 49.43 & 49.97 \\ \midrule
        Compar w/ LB & +6.56 & +7.56 & +8.32 & +9.38 & +10.91 & +11.44  \\ 
        Compar w/ UB & -5.44 & -4.43 & -3.67 & -2.62 & -1.09 & -0.56  \\ \midrule
        DeepLabv3-MobileNetv2 & 36.40 & 38.20 & 39.10 & 39.01 & 40.54 & 40.63 \\ \midrule
        Compar w/ LB & 1.87 & +3.68 & +4.58 & +4.49 & +6.02 & +6.10 \\ 
        Compar w/ UB & -5.44 &-4.43 & -3.67 & -2.62 & -1.09 & -0.56  \\ \bottomrule
    \end{tabular}
    \label{tab:segmentation}
\end{table*}

\begin{table*}[ht]
    \centering
    \caption{Comparing the performances by Mean Average Precision (mAP) of SSD-Lite with MobileNet v2 backbones (up) and Faster R-CNN with ResNet 50 backbones (down) from MiniImageNet and HAID-MiniImageNet. The number in the first line represents the abstract level applied for backbones, from 10 to 1,000 shapes. We also provide the upper bound (UB: fine-tuning with backbone from raster images) and lower bound (LB: training without backbone), and compare them with fine-tuning models with backbones gained from HAID.}
    \begin{tabular}{c|cccccc} \toprule
        ~ & 10 & 30 & 50 & 100 & 500 & 1,000 \\ \midrule
        SSD-Lite & 28.23 & 29.29 & 30.59 & 29.87 & 29.56 & 30.36 \\ \midrule
        Compar w/ LB & +4.34 & +5.40  & +6.70 & +5.98 & +5.66 & +6.47  \\ 
        Compar w/ UB & -2.44 & -1.37 & -0.07 & -0.79 & -1.11 & -0.30  \\ \midrule
        Faster R-CNN & 21.78 & 23.44 & 27.39 & 31.36 & 31.15 & 30.59  \\ \midrule
        Compar w/ LB & +14.97 & +16.63 & +20.59 & +24.56 & +24.35 & 23.79  \\ 
        Compar w/ UB & -5.27 & -3.61 & +0.34 & +4.31 & +4.10 & +3.55  \\ \bottomrule
    \end{tabular}
    \label{tab:object_detection}
\end{table*}

\rev{

\begin{figure*}
    \centering
    \includegraphics[width=\linewidth]{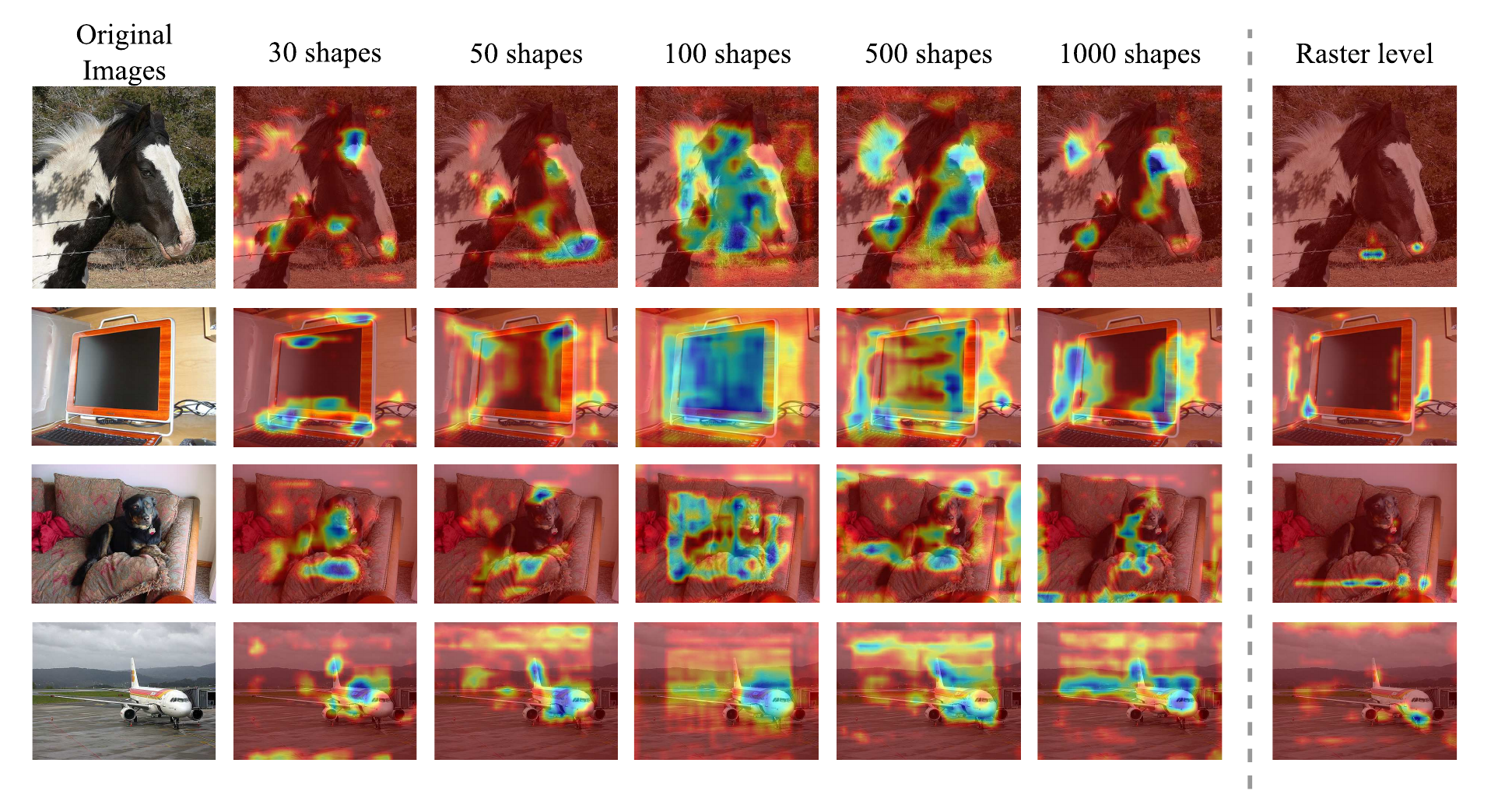}
    \caption{Grad-CAM visualisations from the Faster R-CNN using pretrained backbone. The comparison across the abstract levels of backbones from 30 to 1,000 shapes and original images.}
    \label{fig:cam_faster-rcnn}
\end{figure*}

To further discuss the reason for the surprisingly better results from Faster R-CNN, we apply the Grad-CAM \cite{grad-cam} to visualise the attention of the feature map from Faster R-CNN. The CAM visualisation is shown in the \cref{fig:cam_faster-rcnn}. From the visualisation results, we can observe that the attention maps of Faster R-CNN models with abstract backbones initialised are more tightly concentrated on object geometry and core semantic regions than backbones pre-trained on raster images. As the input detail level of backbones increases to 100 shapes, such geometric focus is the most obvious, which means that it focuses more on the contour or structure representations that benefit box regression, even when fine appearance cues are reduced. Interestingly, this effect weakens as the input detail level of pre-trained backbones increases further (\eg 500 and 1,000 shapes), with attention patterns and localisation performance approaching those of models pre-trained on original raster images. These observations may explain the evaluation results in \cref{tab:object_detection}.

}

\paragraph{Abstraction with all shapes versus triangle only.}
\label{sec:mode0vsmode1}

\begin{figure*}[ht]
    \begin{minipage}[t]{0.5\linewidth}
        \centering
        \includegraphics[width=\linewidth]{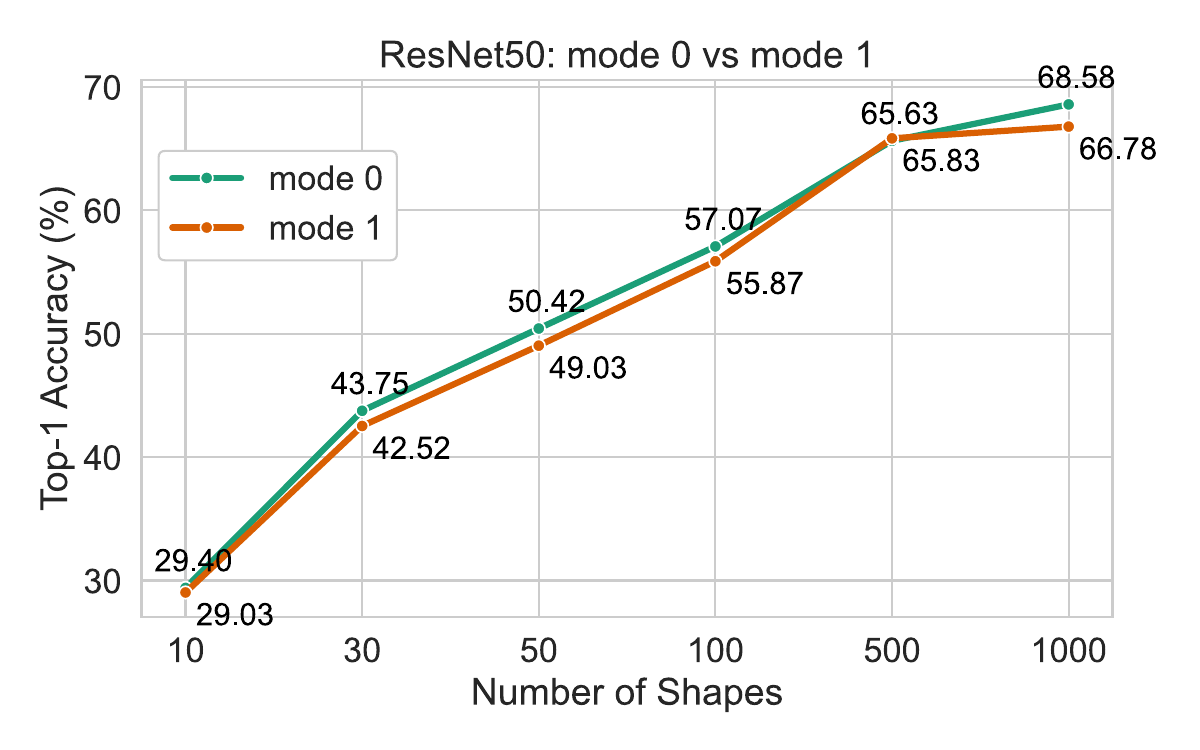}
        \centerline{(a). ResNet50}
        % \label{fig:test_acc_mode0vs1_res}
    \end{minipage}
    \begin{minipage}[t]{0.5\linewidth}
        \centering
        \includegraphics[width=\linewidth]{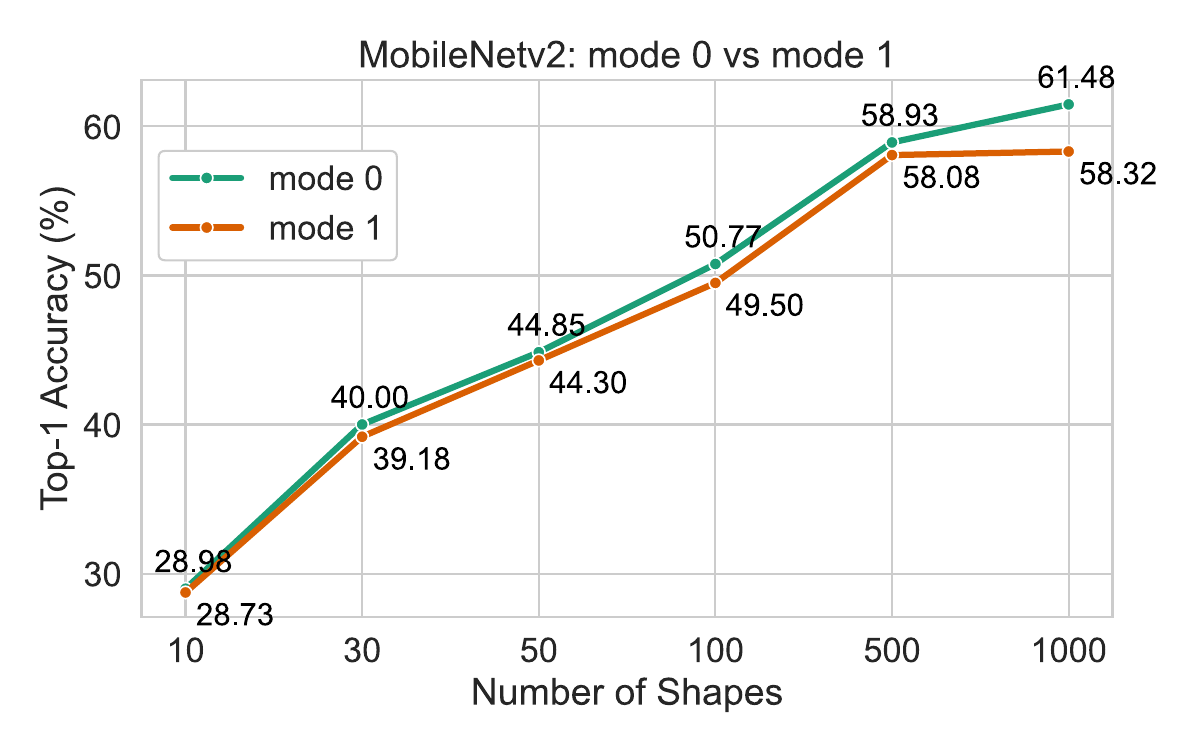}
        \centerline{(b). MobileNet v2}
    \end{minipage}
    \caption{Comparison of ResNet50 and MobileNetv2 performance training on all types of SVG primitives (mode 0) and training on triangular primitives only (mode 1).}\vspace{-2mm}
    \label{fig:test_acc_mode0vs1}
\end{figure*}

In here, we discuss how much difference could be based on the different types of shapes by examining the performance difference between two generation modes: one that employs all available SVG primitives (mode 0) and another that exclusively uses triangles (mode 1). We focus on the triangle-only configuration since it produces images with the lowest file size among the available shape types while maintaining high image quality. \Cref{fig:HAIDexample_img} illustrates the comparisons between mode 0 and mode 1, and more samples can be found in \cref{fig:examples_of_HAID_MiniImageNet}. We trained the networks separately on HAID-MiniImageNet images generated under mode 0 and mode 1 and evaluated them on test sets corresponding to the same levels of abstraction. The performance differences across varying numbers of shapes are presented in \cref{fig:test_acc_mode0vs1}.

\begin{figure*}[ht]
    \centering
    \includegraphics[width=0.8\linewidth]{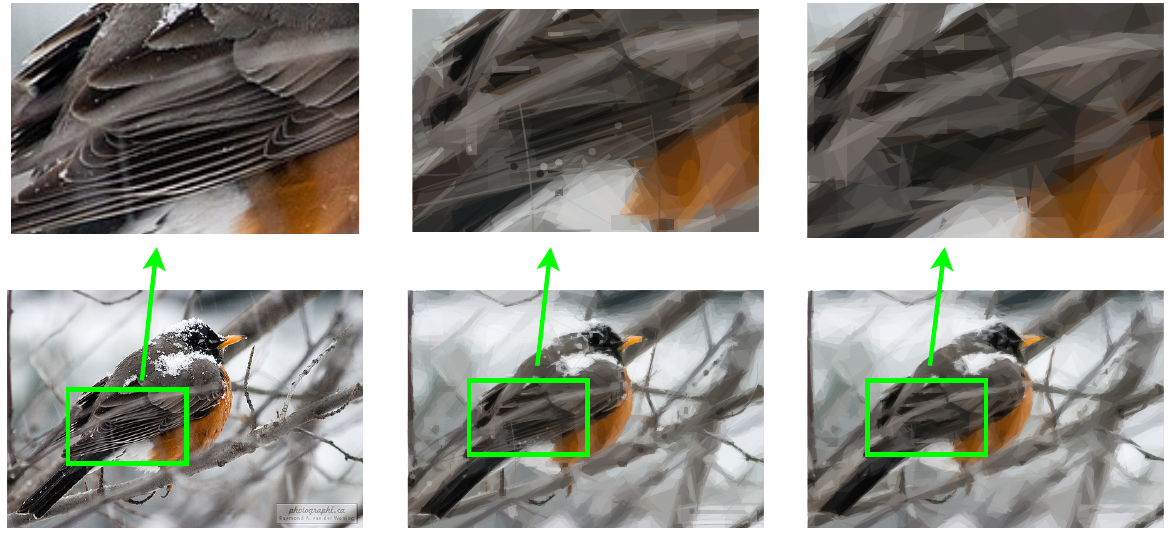}
    \vspace{-2mm}
    \caption{Detailed features comparison: raster image (left), abstract image containing all types of primitive shape (middle), and abstract image containing triangle only (right).}\vspace{-2mm}
    \label{fig:mode0vsmode1}
\end{figure*}

As a result, training on the triangle-only images shows a slight performance drop in most cases. Notably, both ResNet50 and MobileNet v2 exhibit a distinct performance gap between mode 0 and mode 1 when using images with 1,000 shapes; as observed in \cref{fig:mode0vsmode1}, some fine-grained features are not adequately captured in triangle-only images, even at high fine-grained levels. Therefore, considering the advantages of representation performance and displaying details, images with all types of primitive shapes are a priority choice, however, with slight toleration of performance drop, triangle-only images offer a viable alternative with the advantage of lower file capacity cost.

\rev{
\paragraph{User study.}
\label{sec:supp_user_study}

We select 36 images in total across six levels (30, 50, 100, 500, 1,000 shapes, and original images) from HAID-MiniImageNet and MiniImageNet (each level contains 6 images). Images for each abstract level were balanced by a priori difficulty: three single-object images with the simple background (labelled ``easy samples'') and three images with multiple objects, complex textures, or cluttered scenes (labelled ``hard samples''). Participants viewed all 36 images in randomised order and provided a single 1–5 rating per image reflecting how confident they are to perceive the object(s) (1 = cannot recognise at all, 5 = extremely confident). We provide the explicit instruction: \textit{"Your rating should reflect \textbf{how clearly you perceive} the object in each image— you \textbf{do not} need to know the specific name of the object."}, as the task measured perceptual clarity rather than knowledge.

In \cref{sec:user_study}, we discussed the user study to evaluate the dataset from a human perceptive. In here, the comprehensive results are demonstrated in \cref{tab:user_study}. We also recorded the gender information of participants. Out of the 12 responses, 6 recorded gender as male, and 6 recorded gender as female. From the results, there is no significant difference in the rating results between the two genders.

\begin{table*}[ht]
    \centering
    \caption{The table compares the Mean Opinion Score (MOS) from participants of the user study. We also compared the MOS for different genders, which are demonstrated in the rows of `Male' and `Female'.}
    \begin{tabular}{l|ccccc|c} \toprule
    \multirow{2}{*}{Settings} & \multicolumn{5}{c|}{\textbf{Abstract Levels (number of shapes)}}  &   \\ \cmidrule{2-7}
                & 30  & 50  & 100  & 500  & 1,000  & Original images \\ \midrule
    Easy samples   & 2.22      & 3.39      & 3.81       & 4.75       & 4.89        & 4.97            \\
    Hard samples   & 1.81      & 1.5       & 3.11       & 4.44       & 4.67        & 4.94            \\
    All samples    & 2.01      & 2.44      & 3.46       & 4.6        & 4.78        & 4.96            \\  \midrule
    Male    & 2.20      & 2.47      & 3.47       & 4.75       & 4.81        & 4.92            \\
    Female  & 1.84      & 2.42      & 3.44       & 4.45       & 4.75        & 5.00            \\ \bottomrule
    \end{tabular}
    \label{tab:user_study}
\end{table*}

}
\subsection{The correlation between image entropy and abstract images}
\label{sec:entopy}

\begin{figure}[!ht]
    \centering \hspace{-2mm}
    \begin{minipage}[b]{0.46\linewidth}
        \centering
        \includegraphics[width=\textwidth]{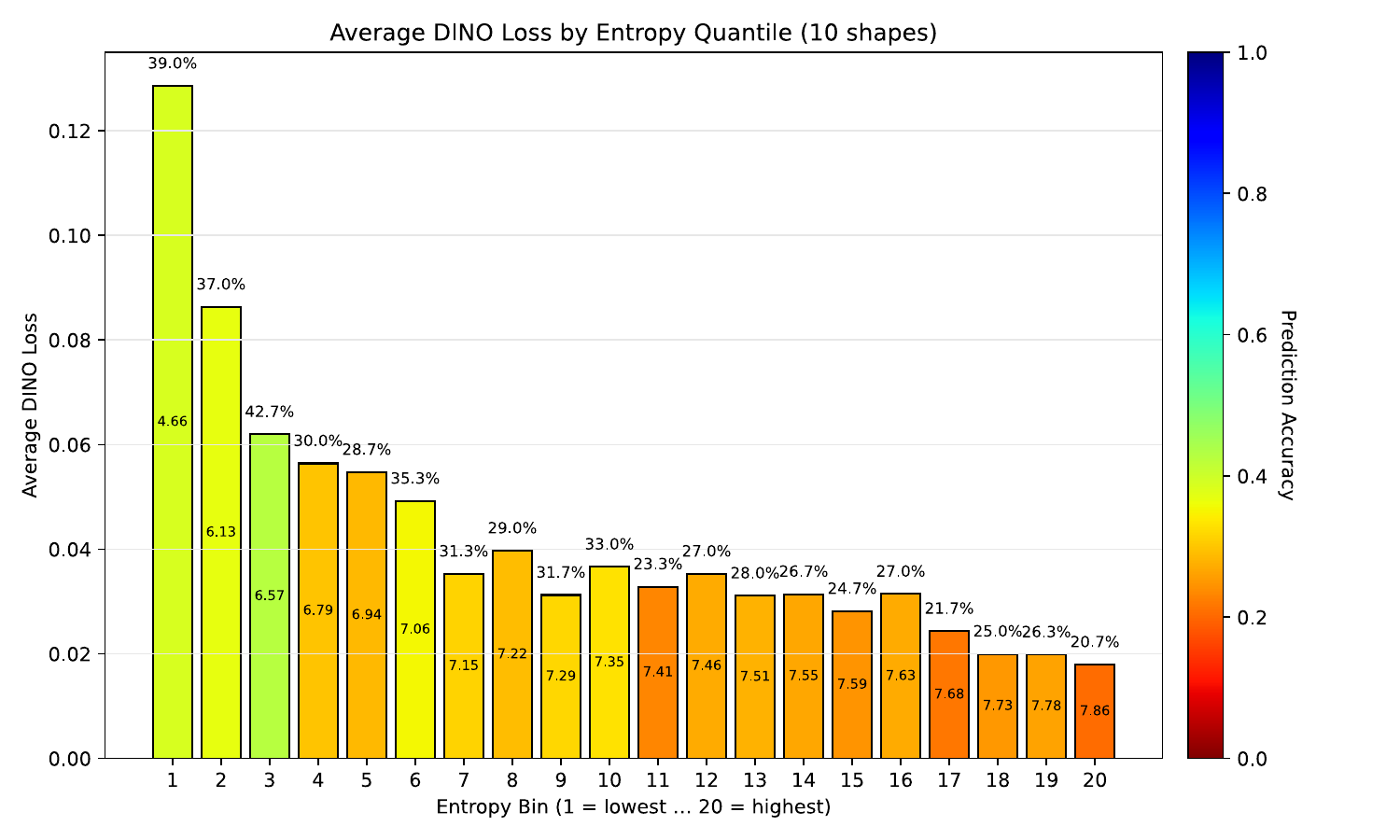}
        \centerline{(a). 10 shapes}
    \end{minipage}\hspace{-2mm}
    \begin{minipage}[b]{0.46\linewidth}
        \centering
        \includegraphics[width=\textwidth]{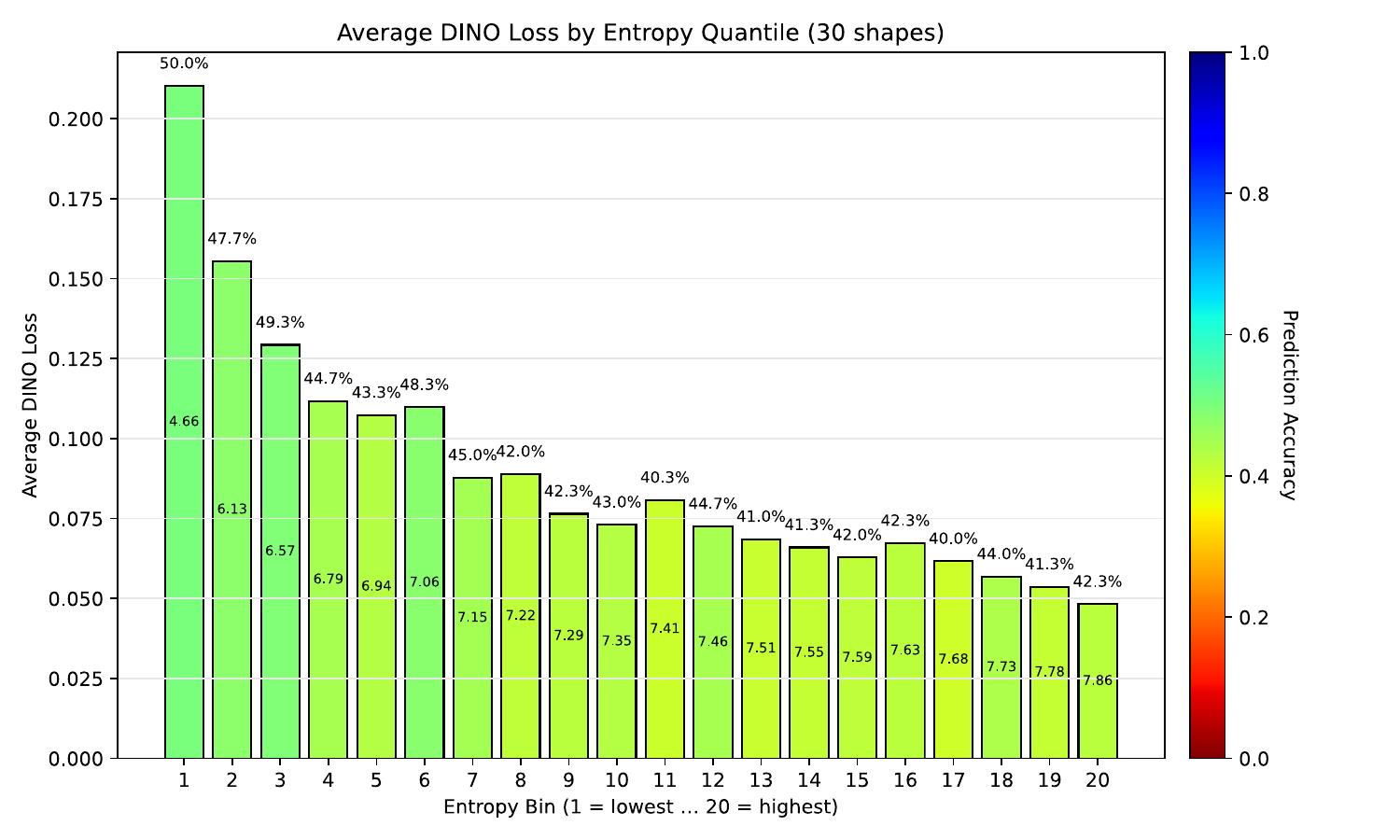}
        \centerline{(b). 30 shapes}
    \end{minipage}\hspace{-2mm}
    \begin{minipage}[b]{0.46\linewidth}
        \centering
        \includegraphics[width=\textwidth]{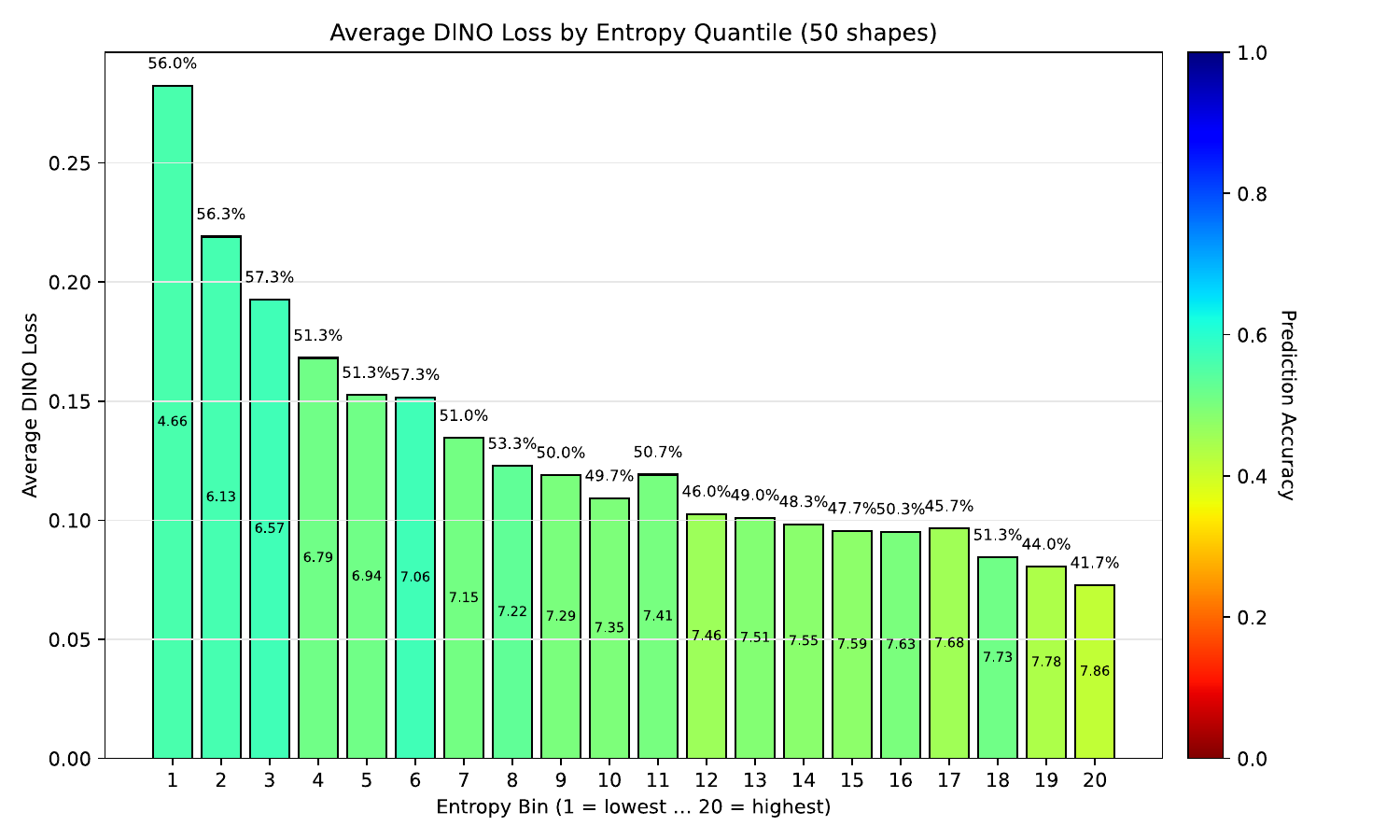}
        \centerline{(c). 50 shapes}
    \end{minipage}\hspace{-2mm}
    \begin{minipage}[b]{0.46\linewidth}
        \centering
        \includegraphics[width=\textwidth]{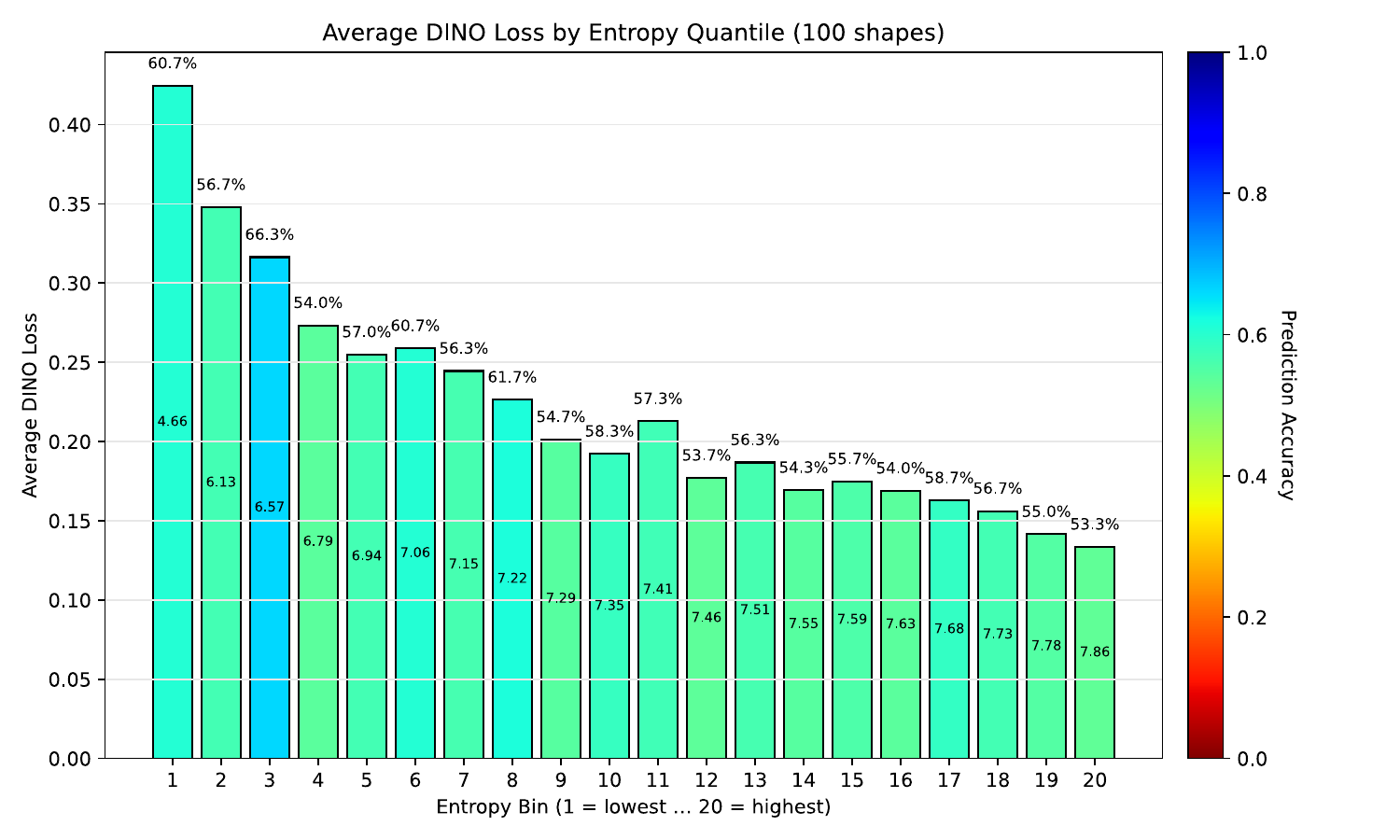}
        \centerline{(d). 100 shapes}
    \end{minipage}\hspace{-2mm}
    \begin{minipage}[b]{0.46\linewidth}
        \centering
        \includegraphics[width=\textwidth]{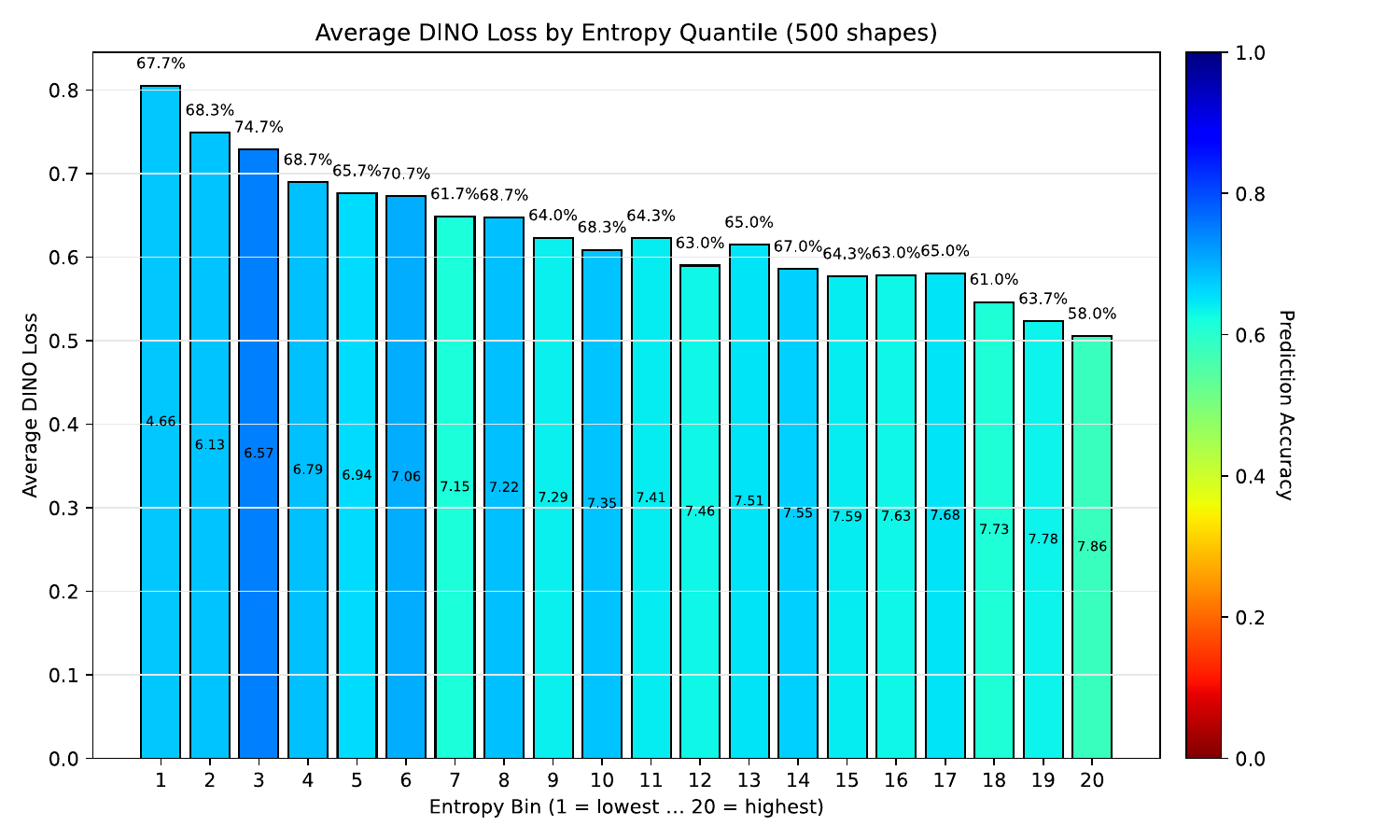}
        \centerline{(e). 500 shapes}
    \end{minipage}\hspace{-2mm}
    \begin{minipage}[b]{0.46\linewidth}
        \centering
        \includegraphics[width=\textwidth]{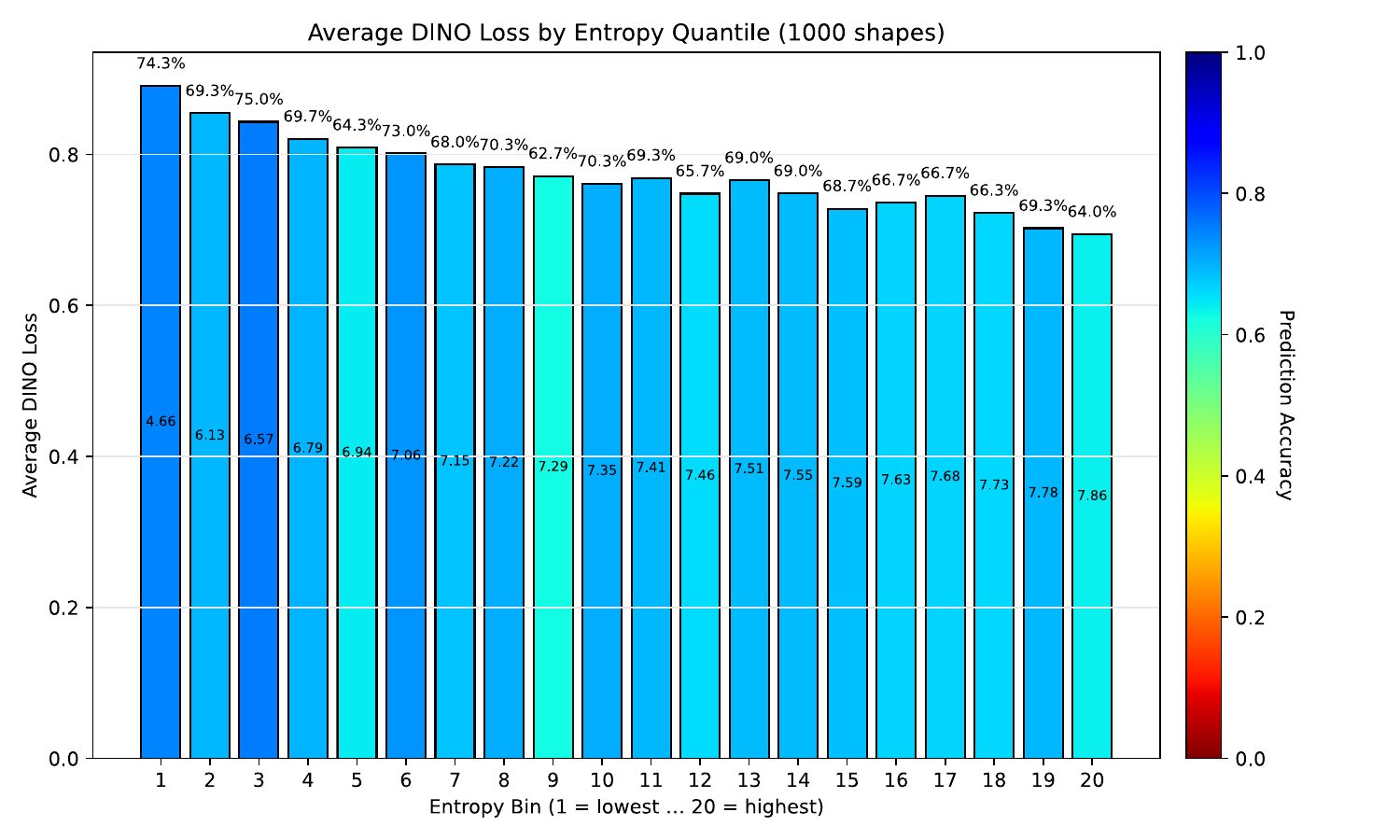}
        \centerline{(f). 1,000 shapes}
    \end{minipage}\hspace{-2mm}
    \caption{The correlation between entropy and DINO loss, each diagram has 20 groups sorted by entropy value from small to large. The colour and the value above each bin represent the prediction accuracy of each bin, and the value inside each bin represents the average entropy value of each group.}
    \label{fig:entopyvsDINO}
\end{figure}

\begin{figure*}[!ht]
    \centering
    \includegraphics[width=\linewidth]{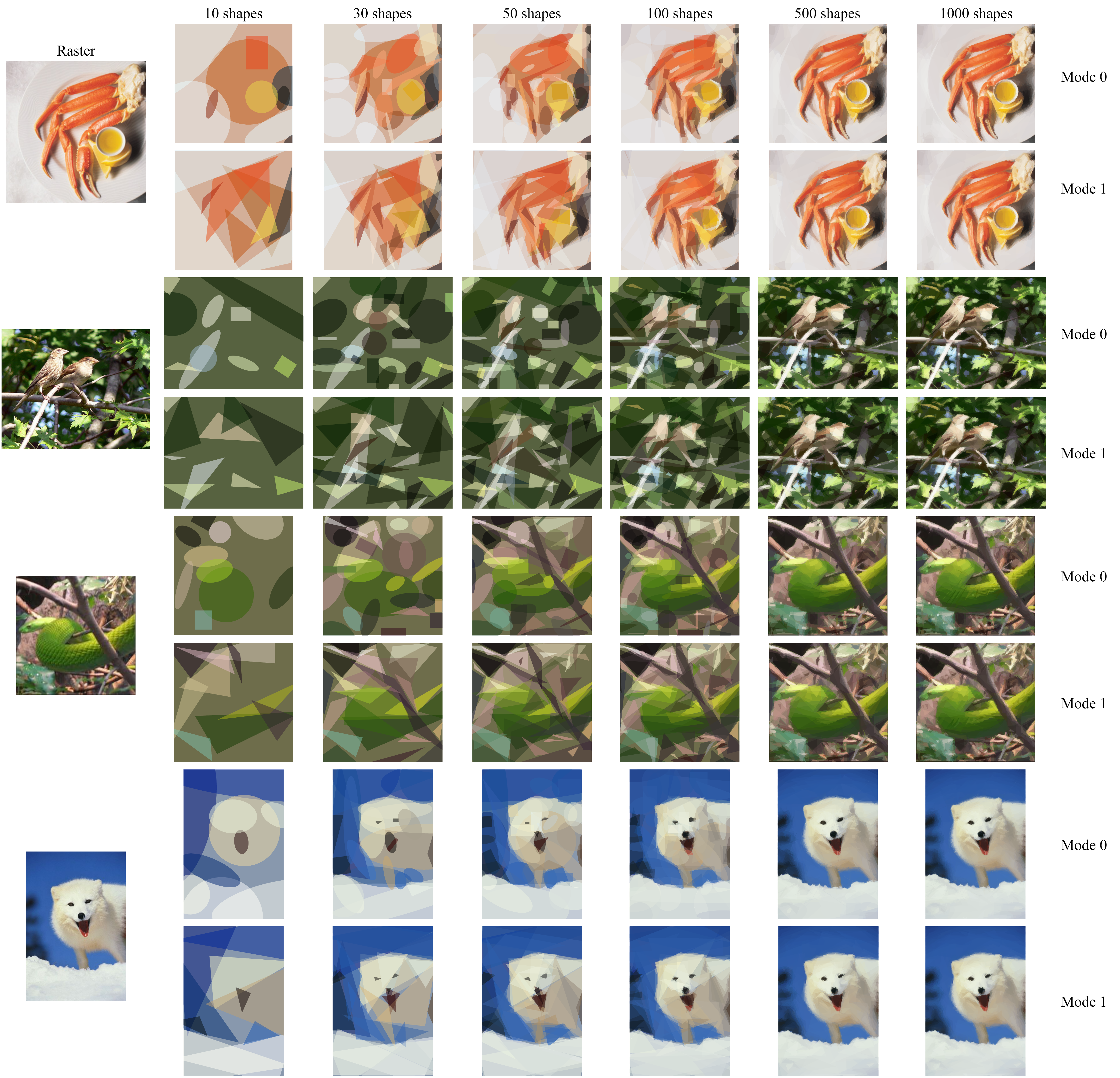}
    \caption{More examples of HAID-MiniImageNet}
    \label{fig:examples_of_HAID_MiniImageNet}
\end{figure*}

Entropy of the image serves as a quantitative measure of its information content as well as its complexity. We observed that, in the human perceptive, for the simple images (\eg with single object and unsophisticated background), such as the last row images in \cref{fig:examples_of_HAID_MiniImageNet}, are generally recognisable with a lower number of shapes than the complex images, such as the second row images in \cref{fig:examples_of_HAID_MiniImageNet}. Therefore, we speculate there is a correlation between the information complexity of the original image and the levels of its abstractions. To prove this hypothesis, we randomly sampled 4000 image pairs from the MiniImageNet and HAID-MiniImageNet, calculating the entropy values of these 4000 images, and sorted them into 20 groups based on the entropy from smallest to largest. We also labelled each group with the perceptual loss (DINO loss specifically) between abstract and original images, their prediction accuracy, and the mean entropy value. The results are shown in \cref{fig:entopyvsDINO}, 

\section{Primitive}
\label{sec:smPrimitive}

\begin{figure}[ht]
    \centering
    \includegraphics[width=\linewidth]{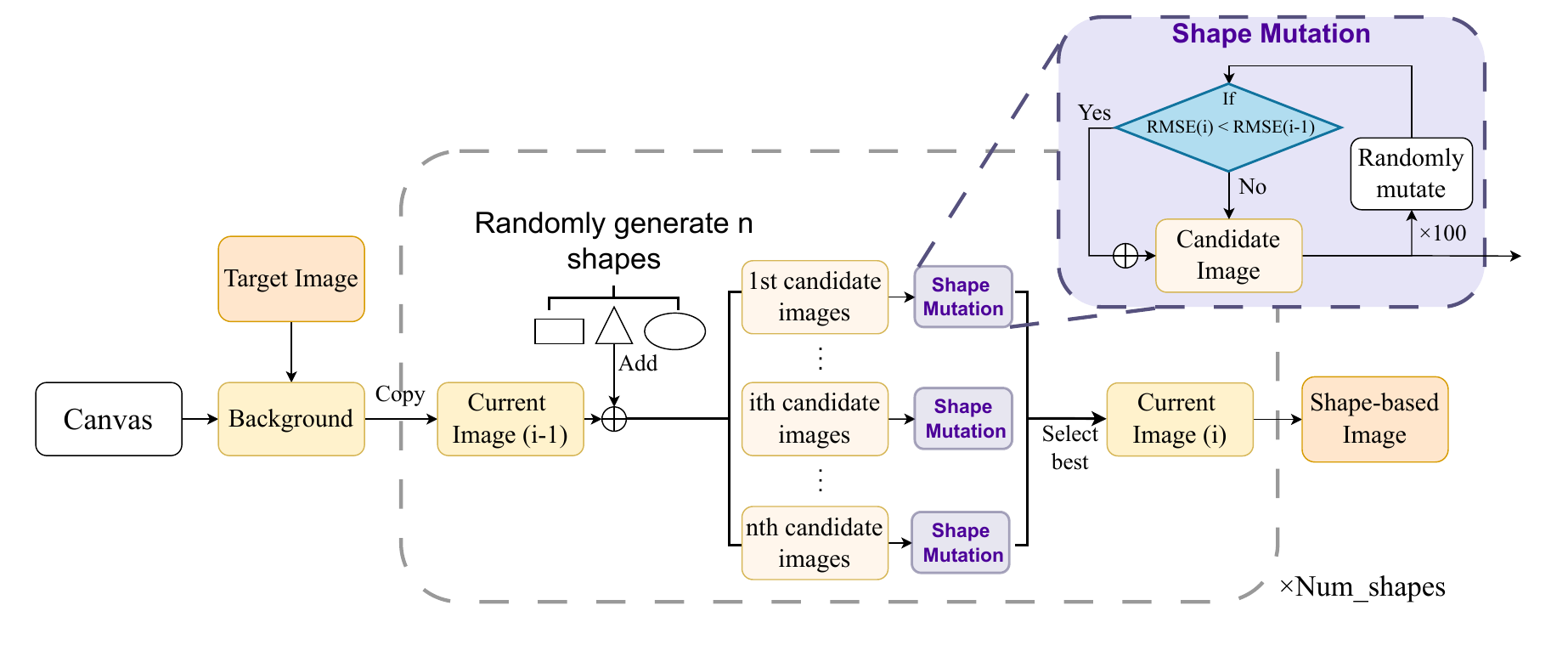}
    \caption{The generating procedure of Primitive.}
    \label{fig:Primitive_procedure}
\end{figure}

The \cref{fig:Primitive_procedure} specifically explained how the Primitive generates the shape-based images from the provided raster images. 

\end{document}